\documentclass[preprint]{elsarticle}

\usepackage{amssymb}
\usepackage{amsmath}
\usepackage[utf8]{inputenc}
\usepackage[T1]{fontenc}
\usepackage{graphicx}
\usepackage{booktabs}
\usepackage{array}
\usepackage{subcaption}        
\usepackage{siunitx}           
\usepackage{url}
\usepackage{lineno}
\usepackage[hidelinks]{hyperref}

\journal{Acta Astronautica (preprint)}

\begin{document}

\begin{frontmatter}


\title{RoverDevKit: An open, physics-grounded tradespace toolkit for conceptual design of lunar micro-rovers}

\author[duke]{Jon Reifschneider \corref{cor1}}
\ead{jon.reifschneider@duke.edu}
\cortext[cor1]{Corresponding author}

\affiliation[duke]{organization={Duke University},
            city={Durham},
            state={NC},
            postcode={27708}, 
            country={USA}}

\begin{abstract}
The number of planned lunar micro-rovers is increasing, yet few open, benchmarked tools exist to explore trade-offs in their design. Rover sizing is a highly coupled problem with a feedback loop between vehicle mass and mobility: a heavier rover sinks deeper into loose regolith, deeper sinkage increases power demand, more power requires a larger array, and the array adds mass. We present RoverDevKit, an open-source tradespace exploration toolkit with an evaluator that couples a closed-form wheel--soil model with bottom-up mass, solar-power, thermal-survival, and traverse models. The evaluator is fast enough to serve as the fitness function of a multi-objective genetic algorithm (NSGA-II): a search comprising $3\,000$ mission evaluations takes under a minute on a single laptop core. We use it to map trade-offs among distance traveled, vehicle mass, and slope capability on mare, polar, highland, and crater-rim missions. The limiting factor varies by mission: energy storage at high latitude, slope traction on loose highland regolith, and traverse range on mare and crater-rim terrain. The wheel--soil model reproduces measured single-wheel drawbar pull from two independent datasets with accuracy consistent with the $20$--$30\,\%$ error typically reported for this class of model, and the mass model predicts published $5$--$50$\,kg rover masses with median absolute percentage error of $10.5\,\%$. A comparison with five published rover designs in the same mass class shows that they lie near the predicted Pareto front.
\end{abstract}


\end{frontmatter}



\section{Introduction}
\label{sec:intro}

The cadence of lunar surface exploration has accelerated over the past decade, driven by growing national lunar science campaigns, the rise of commercial landers, and public-private initiatives such as NASA's Commercial Lunar Payload Services program. As the number of rovers on the surface has increased, their size has decreased with the application of new technologies, and micro-rovers weighing under 50\,kg have become a rapidly growing vehicle class for lunar surface exploration. Example rovers in this class include Pragyan and Rashid-1 at roughly 26\,kg and 10\,kg respectively \cite{Chandrayaan3EoPortal,alzaabiOperationalConceptsRehearsal2024}, along with a new generation of even smaller designs including the Cooperative Autonomous Distributed Robotic Exploration (CADRE) swarm units from NASA \cite{jpl_cadre} and iSpace's Tenacious rover \cite{IspaceEUROPEAnnouncesCompletion}. However, while the number of rovers developed has grown, the availability of shared tooling for the conceptual design of these vehicles has not kept pace. Integrated early-stage design capability remains concentrated in proprietary concurrent-engineering facilities available mainly to government agencies and large corporations \cite{caseEvolutionTeamX252021}, while the published design reports of the smaller teams that have produced much of this vehicle class use requirements decomposition and iterative prototyping of a point design rather than automated search over a coupled design space \cite{dellatorreAMALIAMissionLunar2010,walkerUpdateQualificationHakuto2016,caraccioARDITOModularTechnology2025}, as Section~\ref{sec:related-work} discusses. Open robot simulators are mature and widely used, but their use centers on simulating an already-defined vehicle rather than sizing one via optimization. The lunar micro-rover class, despite its growth and criticality to the plans of national space agencies and private corporations, is therefore not well served by open conceptual design tooling that couples mobility, power, thermal, and mass sizing with public empirical benchmarking and reproducible artifacts. 

Lunar micro-rover subsystems are strongly interdependent, as they are on most spacecraft. What is specific to a surface rover is the role of the terrain: a heavier rover sinks deeper into loose regolith, deeper sinkage takes more power to drive through, more power requires a larger solar array, and that array and its supporting structure add further mass. Vehicle mass therefore feeds back on itself. Launch vehicle and lander allocations cap the mass and volume available to absorb that feedback, and at micro-rover scale the fixed overheads of motors, avionics, and structure leave little margin within which to resolve it. This regime lends itself to coupled tradespace evaluation rather than decoupled subsystem analyses and manual trades.

Beyond the highly coupled nature of rover subsystems, additional factors make rover design particularly challenging. Mobility failures of rovers on the lunar surface are unrecoverable and therefore the characterization of the feasible region is particularly critical. Because soil and solar potential vary widely by location, the feasible design region is driven largely by the mission profile: an optimized 10\,kg rover used for polar prospecting will be vastly different from a similarly sized rover for lunar mare operations. Public design data is scarce: full design vectors could be reconstructed for only five in-class rovers (Section~\ref{sec:rediscovery}), too few to induce class-specific design rules from flown vehicles alone. Heuristics used on heavier planetary rovers are not guaranteed to apply here either, and Section~\ref{sec:design-rules} gives a case where one does not: the expectation that six-wheel rocker-bogie layouts become optimal as mass grows does not hold within this class.

Closing this tool gap requires models that are both credible and fast enough for early-stage tradespace exploration. Terramechanics is the central challenge: it often dominates both uncertainty and runtime, yet it must be evaluated repeatedly inside optimization loops. Coupling terramechanics to energy, mass, and thermal models closes the design loop, but it also raises the cost of each candidate evaluation.

This work presents RoverDevKit, an open and empirically benchmarked tradespace toolkit for quantifying how lunar mission profiles reshape the feasible range--mass--slope envelope of sub-$50$\,kg rovers. The paper makes three contributions.
\begin{enumerate}
\item \emph{An open mission evaluator} spanning terramechanics, mobility, power, thermal, mass, and traverse, benchmarked against measured data and fast enough to serve directly as the fitness function of a multi-objective optimizer: one Pareto front of $3\,000$ mission evaluations costs under a minute on a single core (Section~\ref{sec:runtime}).
\item \emph{Layered empirical checks} that state what each layer can and cannot show: component comparisons against measured single-wheel data and against published rover masses, consistency checks against two flown rovers, and a system-level check of whether the optimizer selects designs near five real rovers. We then propagate the wheel--soil comparison error through the optimizer to test whether the design conclusions survive it. These are standard verification practices; the contribution is applying and reporting them openly for a class where such benchmarking is rarely published.
\item \emph{Mission-dependent design rules} from NSGA-II Pareto fronts across four canonical missions. The limiting factor changes with the mission: energy storage at high latitude, slope traction on loose highland regolith, and traverse range on mare and crater-rim missions. Four-wheel layouts Pareto-dominate the modeled range--mass--slope envelope, overturning the expectation (including our own pre-study hypothesis) that the rocker-bogie six-wheel option becomes optimal at the heavy end.
\end{enumerate}

\begin{figure}[htbp]
  \centering
  \includegraphics[width=\linewidth]{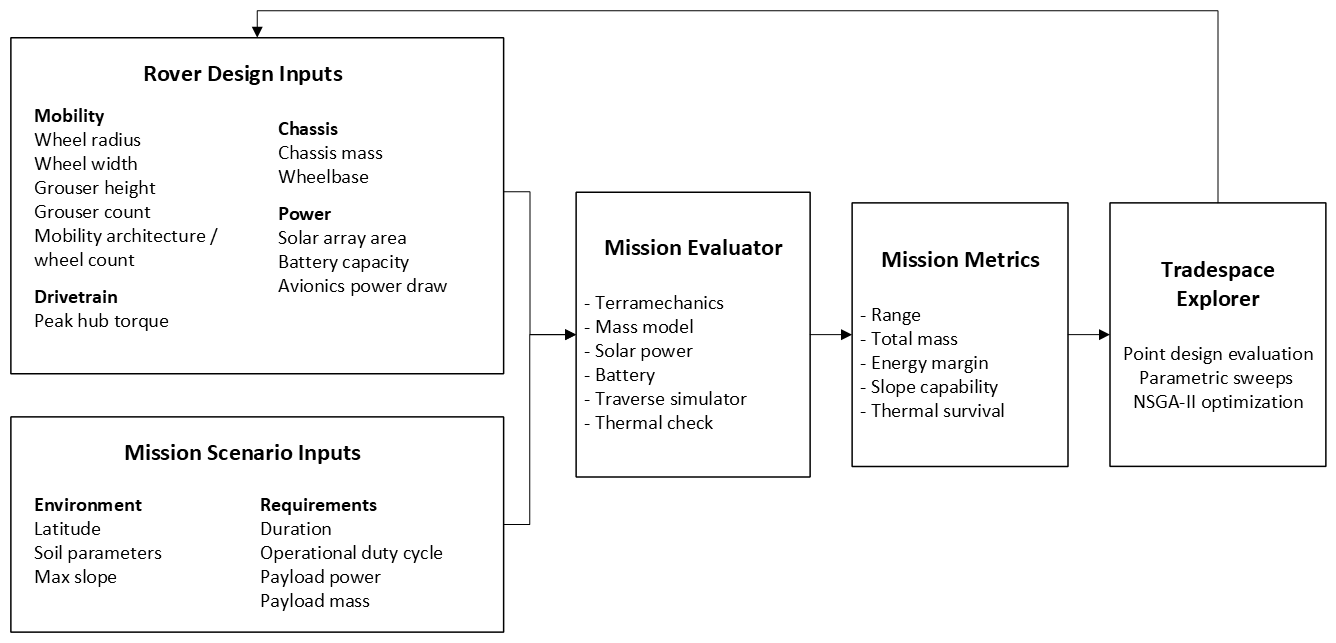}
  \caption{System architecture of RoverDevKit. A design vector and mission scenario enter the mission evaluator, which couples terramechanics, mass, power, thermal, and traverse sub-models to produce range, mass, energy, and slope-capability metrics. Those metrics are consumed by the tradespace explorer for point-design evaluation, parametric sweeps, and NSGA-II optimization.}
  \label{fig:workflow}
\end{figure}

Figure~\ref{fig:workflow} summarizes this architecture. The modeling approaches used in RoverDevKit follow established methods from aerospace conceptual design and multi-objective optimization; the contribution is not the optimization algorithm itself but the open, benchmarked evaluator, the evidence assembled for it, and the mission-dependent design rules it exposes.

\paragraph{Terminology.}
The \emph{evaluator} is the coupled analysis chain that maps one design vector and mission scenario to performance metrics. Its \emph{sub-models} are the five physics components that chain couples---terramechanics, mass, power, thermal, and traverse---each referred to individually as a model, as in the mass model. The \emph{kernel} is the wheel--soil contact solve inside the terramechanics sub-model.

\section{Background and related work}
\label{sec:background}

\subsection{Background}
\label{sec:background-theory}

\subsubsection{Conceptual design and the tradespace-exploration paradigm}
During the concept development stage of system development, the functional requirements of a system are mapped to a physical form and a design concept is selected after an analysis of alternatives process in which a subset of point designs from the tradespace are selected and explored. The selection of a design concept during conceptual development typically locks in a majority of the cost and schedule of the system development \cite{rossTradespaceExplorationParadigm2005}. It is therefore critical to explore the tradespace as thoroughly as possible during this phase of development to understand the impact of requirements and constraints on the feasible region and to select an optimal design concept. However, manual analysis of design candidates is slow and costly, limiting the number of candidate designs that teams can explore in practice. Automated evaluation pipelines can accelerate this process and enable teams to run parameter sweeps and optimization loops to rapidly characterize the design space. The cost of a single evaluation is what limits how much of the space can be searched this way. Where that cost is high, the established remedy is surrogate-assisted optimization, in which a data-driven model fitted to samples of the expensive evaluator stands in for the fitness function \cite{jinSurrogateassistedEvolutionaryComputation2011,forresterRecentAdvancesSurrogate2009}. This paper takes the other route, keeping the physics-based evaluator itself cheap enough to serve as the fitness function directly. The lunar micro-rover regime lacks open, empirically benchmarked tooling for this purpose, and this is the accessibility gap that RoverDevKit fills.

\subsubsection{Terramechanics models}
Wheel terramechanics modeling is typically a limiting factor in the accuracy of modeling approaches for lunar and planetary rovers as well as the slowest step in an evaluation pipeline for design concepts. Reviews of the field classify wheel--soil models into three frameworks: empirical, semi-empirical, and physics-based numerical \cite{taheriTechnicalSurveyTerramechanics2015,swamyReviewModelingValidation2023}. Only the latter two are useful for conceptual design, because empirical mobility indices predict performance from aggregate vehicle properties rather than computing it from the wheel dimensions that this paper treats as design variables. The next two paragraphs describe them in order of increasing fidelity and cost. A third paragraph then describes the experimental single-wheel testbeds that supply the ground truth for validating both.

\paragraph{Classical semi-empirical models.}
The foundational analytical framework is based on Bekker's pressure--sinkage relation \cite{bekkerIntroductionTerrainvehicleSystems1969} combined with the Wong--Reece model of wheel--soil stresses \cite{wongPredictionRigidWheel1967}. For a rigid wheel, the normal pressure $p$ that the terrain supports grows with sinkage $z$ as per equation \eqref{eq:bekker} below
\begin{equation}
  p(z) = \left(\frac{k_c}{b} + k_\phi\right) z^{\,n},
  \label{eq:bekker}
\end{equation}
where $b$ is the smaller dimension of the contact patch, $n$ is the sinkage exponent, and $k_c$ and $k_\phi$ are the cohesive and frictional moduli of the soil \cite{bekkerIntroductionTerrainvehicleSystems1969}. Integrating this pressure over the contact patch, out to the equilibrium sinkage at which the terrain supports the applied load, yields the compaction resistance: the force equivalent of the work per unit travel distance spent compacting regolith ahead of the wheel. Tangential traction is generated by soil shear, and the shear stress $\tau$ follows the Mohr--Coulomb failure criterion combined with the Janosi--Hanamoto shear-displacement law,
\begin{equation}
  \tau(j) = \bigl(c + \sigma \tan\phi\bigr)\!\left(1 - e^{-j/K}\right),
  \label{eq:janosi}
\end{equation}
where $c$ is the soil cohesion, $\phi$ the internal friction angle, $\sigma$ the local normal stress, $j$ the soil shear displacement, and $K$ the shear-deformation modulus \cite{janosi1961}. Shear displacement is driven by wheel slip per equation \eqref{eq:slip} below.
\begin{equation}
  s = 1 - \frac{v}{\omega r},
  \label{eq:slip}
\end{equation}
for forward velocity $v$, angular rate $\omega$, and wheel radius $r$. Integrating $\tau$ over the contact patch yields the gross tractive effort. Classically, the net drawbar pull available to accelerate the vehicle and climb slopes is then taken as that gross tractive effort less the motion resistance, of which compaction resistance is the dominant term. Because these equations require only a handful of measured soil parameters and evaluate quickly they are usable within an optimization loop, at the expense of some potential accuracy loss due to the assumptions of a rigid wheel and steady-state operation. The same semi-empirical laws can also be applied over a discretized terrain surface instead of a single contact plane, as in the Soil Contact Model (SCM), which adds three-dimensional contact patches, plastic soil deformation, bulldozing, and multi-pass effects \cite{krennSCMSoilContact2009,serbanRealTimeSimulationGround2023}.

\paragraph{Higher-fidelity numerical models.}
Physics-based numerical methods discretize the soil itself, either as a continuum solved by finite-element or meshless schemes such as the material point method or smoothed-particle hydrodynamics \cite{huTractionControlDesign2022}, or as individual grains tracked by the discrete-element method (DEM) \cite{recueroHighfidelityApproachVehicle2017}. These resolve grouser-scale granular flow, deformable-wheel coupling, and transient maneuvers that the closed-form kernel cannot represent, and because their inputs are physical soil properties they need little of the calibration the semi-empirical models require. The cost is steep: reported real-time factors are as low as $30$--$150$ for continuum methods and in the range $3\,000$--$15\,000$ for DEM \cite{huTractionControlDesign2022,recueroHighfidelityApproachVehicle2017}, so a single wheel--soil run consumes hours of wall-clock time. That expense is bearable for focused verification of a chosen design but not for a search over tens of thousands of candidates.

\paragraph{Experimental single-wheel testbeds.}
Controlled single-wheel experiments on lunar soil simulants provide the ground truth used to validate the analytical approaches. Experimental data from studies used to benchmark our kernel include Ding et al. \cite{dingExperimentalStudyAnalysis2011} (smooth and grousered wheels on lunar simulant), Wang \& Han \cite{wangcheng-canExperimentalStudyLunar2016} (KLS-1 simulant), and Hurrell et al. \cite{hurrellTractionPerformanceEvaluation2025} (Rashid-1 wheel on FJS-1). 

\subsection{Multi-objective optimization}
The design of a lunar rover is inherently a multi-objective problem: increasing range or slope capability tends to increase wheel, battery, solar array and structural mass. Rather than combining these trade-offs into a single utility function, we use Pareto dominance to evaluate designs. For a set of $m$ objectives $f_1,\dots,f_m$, all expressed in minimization form, design $a$ dominates design $b$ if
\begin{equation}
\begin{aligned}
  f_k(a) &\le f_k(b) && \forall k \in \{1,\dots,m\}, \\
  f_j(a) &< f_j(b) && \text{for at least one } j \in \{1,\dots,m\}.
\end{aligned}
\label{eq:pareto-dominance}
\end{equation}
The set of non-dominated points forms the Pareto front, or the set of designs for which improving one objective requires sacrificing another. A scalar heuristic (a weighted sum, simulated annealing, or standard particle-swarm search) would collapse that front to a single point and hide the result this paper is written to show: that the limiting objective among range, mass, and slope capability changes with mission profile. Gradient-based methods are a poor fit for the same reason and because the design space is mixed-variable and non-convex (discrete architecture, wheel count, and grouser count). We therefore use a Pareto-based evolutionary algorithm. NSGA-II \cite{debFastElitistMultiobjective2002} is a standard choice for this class of problem: it ranks candidates by Pareto dominance and keeps a spread of designs along the front rather than collapsing to a cluster. Because the evaluator is already fast enough to serve as the fitness function (Section~\ref{sec:runtime}), the published fronts are computed against it directly, with no surrogate approximation needed in the loop.

\subsection{Related work}
\label{sec:related-work}
The need for rapid, coupled trade studies in the design of space systems is well recognized, but existing approaches do not yet provide an open, empirically benchmarked tradespace tool for lunar micro-rovers. Major concurrent-engineering facilities such as NASA's Jet Propulsion Laboratory (JPL) Team-X \cite{caseEvolutionTeamX252021} and the European Space Agency (ESA) Concurrent Design Facility establish the value of integrated Pre-Phase-A analysis, but their models, databases, and workflows are generally not released in a form that academic or smaller industry teams can inspect and extend. Rover-specific tools have also been developed, including the MIT Mars Surface Exploration tool for rover mission concepts \cite{lamamyDesigningNextGeneration2006}, the German Aerospace Center (DLR) RODENT tool for preliminary lunar and Martian rover sizing \cite{jagulaModuleBasedDesign2008}, ESA's Rover Chassis Evaluation Tool for locomotion and chassis evaluation \cite{michaudRoverChassisEvaluation2006}, the computational planetary-rover sizing and optimization tool of Seeni and Sch\"afer \cite{seeniComputationalToolConceptual2023}, and broader planetary-robotics tools such as MISTRAL for sampling systems \cite{riccobonoMultidisciplinaryDesignTool2021}. More recent studies apply multidisciplinary optimization or model-based systems engineering (MBSE) simulation to solar-powered and lunar rover concepts \cite{kimMultiDisciplinaryMultiObjective2022,alhajeriModelBasedSystems2023}. These establish that coupled rover sizing is tractable, but each addresses a different scope and they are reported in the literature rather than distributed as software. None is available as an open, benchmarked evaluator for the sub-$50$\,kg lunar class.

How design is actually carried out in this vehicle class is documented by the competition and commercial programs that have produced much of it. Team Italia's AMALIA rover, a Google Lunar X Prize candidate, reports a subsystem-level conceptual design shaped by the cost and schedule constraints of a commercially funded mission rather than an institutional one \cite{dellatorreAMALIAMissionLunar2010}. The Hakuto micro-rover, another Google Lunar X Prize entrant in this size class, documents an architecture reached by iterating several prototype phases and then requalifying commercial off-the-shelf parts for flight \cite{walkerUpdateQualificationHakuto2016}. Among European Rover Challenge teams, the ARDITO platform documents an ECSS-derived requirements and functional analysis feeding a product-tree decomposition across subsystems \cite{caraccioARDITOModularTechnology2025}. These are rigorous, well-documented processes, but they converge on a point design through requirements decomposition and physical iteration; none reports a coupled mobility--power--thermal--mass evaluator used to search a design space. The gap this paper targets is therefore not engineering rigor in this community but the absence of shared, open tooling for coupled early-stage trades.

A separate and more mature class of open tools exists for robot simulation, including Gazebo \cite{koenigDesignUseParadigms2004}, Webots \cite{michelWebotsProfessionalMobile2004}, and CoppeliaSim \cite{rohmerVREPVersatileScalable2013}. These simulate the dynamics, control, and perception of an already-defined vehicle. Conceptual design is the inverse problem: the vehicle is the output, so the mass, power, and thermal models of Sections~\ref{sec:mass-model} and~\ref{sec:power-thermal} would still have to be built. A simulator would replace only the traverse loop of Section~\ref{sec:traverse}, and two mismatches remain. One Pareto front here costs thousands of multi-week mission evaluations (Section~\ref{sec:runtime}), far above the budget of a time-stepped rigid-body simulator. Their default contact models also treat wheel--ground interaction as rigid-body friction rather than sinkage-dependent pressure and shear in deformable regolith (Section~\ref{sec:background-theory}). These tools are the right choice for detailed design, control, and autonomy once a concept has been selected.

RoverDevKit addresses the remaining gap: an open, empirically checked tradespace evaluator tailored to the sub-$50$\,kg lunar micro-rover regime. It combines a released design dataset and evaluator, a wheel--soil kernel benchmarked against measured single-wheel experiments, a bottom-up mass model checked against published rover masses, and Pareto-front analyses that map the feasible range--mass--slope envelope for representative mission classes. Its computational approach follows established aerospace design practice, in which evolutionary multi-objective optimization is a mature approach to coupled, mixed-variable tradespaces \cite{debFastElitistMultiobjective2002}. The contribution made here is a reproducible and benchmarked design tool for a rapidly growing rover class. The paper shows that changing mission profile within the same sub-$50$\,kg envelope can shift the active design trade from energy storage to slope traction to traverse range. 

\section{Mission evaluator}
\label{sec:evaluator}

\subsection{Design vector and mission scenario}
\label{sec:design-vector}
The physics-based evaluator maps a single rover design and mission scenario to a set of mission-level metrics. The input design vector contains the set of variables representing a rover design which can be directly traded: wheel geometry, mobility architecture, chassis structural mass, wheelbase, solar array area, battery capacity, avionics power, and peak per-wheel hub torque. Architecture is restricted to a rigid four-wheel layout or a rocker-bogie six-wheel layout and fixes the wheel count; no other six-wheel arrangement is in the search. The mission scenario contains the environmental and operational requirements for the rover given its mission: latitude, soil family, maximum slope, mission duration, operational duty cycle, and scientific payload mass and power. Payload is considered in the scenario rather than the design vector because it is a mission requirement specified by the science objectives, while the rover bus is designed around it. One entry, wheelbase, is included and reported but is not used in any sub-model in the present implementation: it would enter through load distribution and tip-over stability, neither of which is modeled (Section~\ref{sec:traverse}). It is retained because it is needed to describe a design and to compare against published rovers, but the optimizer has no signal on it, so its values on the reported fronts should not be read as recommendations. 

The four mission scenario families in Table \ref{tab:design-ranges} are representative cases chosen to isolate distinct drivers in the lunar micro-rover regime: moderate-latitude mare traversal, low-sun polar prospecting, loose-regolith highland slope capability, and short crater-rim survey missions. The continuous scenario variables are varied within each family so the resulting design rules reflect mission-regime effects rather than a single hand-picked operating point.

\begin{table}[htbp]
  \centering
  \small
  \setlength{\tabcolsep}{4pt}
    \caption{Design variable and scenario ranges. \emph{Top:} the 11-D design vector the optimizer trades. \emph{Bottom:} the four canonical mission scenario families.}
  \label{tab:design-ranges}
  \begin{tabular}{llcc}
    \toprule
    \multicolumn{4}{l}{\textbf{Design variables}} \\
    \midrule
    Variable & Symbol & Range & Units \\
    \midrule
    Wheel radius              & $r$      & $0.05$--$0.20$ & m \\
    Wheel width               & $b$      & $0.03$--$0.20$ & m \\
    Grouser height            & $h_g$    & $0$--$0.020$   & m \\
    Grouser count             & $N_g$    & $0$--$24$      & -- \\
    Mobility arch.\ / wheel count & $N_w$ & rigid 4 / rocker 6 & -- \\
    Chassis (structural) mass & $m_c$    & $0.5$--$50.0$  & kg \\
    Wheelbase                 & $L_{\mathrm{wb}}$ & $0.3$--$1.2$ & m \\
    Solar-array area          & $A_s$    & $0.1$--$1.5$   & m$^2$ \\
    Battery capacity          & $C_b$    & $5$--$500$     & Wh \\
    Avionics power            & $P_a$    & $5$--$40$      & W \\
    Peak hub torque           & $T_{\mathrm{hub}}^{\mathrm{peak}}$ & $0.05$--$20.0$ & N\,m \\
    \midrule
    \multicolumn{4}{l}{\textbf{Mission scenario families}} \\
    \midrule
    Scenario family                    & Latitude (deg) & Duration (Earth d) & Max slope (deg) \\
    \midrule
    Equatorial mare traverse  & $10$ to $25$   & $10$--$18$ & $3$--$18$ \\
    Polar prospecting         & $-88$ to $-80$ & $25$--$35$ & $10$--$28$ \\
    Highland slope capability & $5$ to $20$    & $5$--$10$  & $18$--$30$ \\
    Crater-rim survey         & $-15$ to $15$  & $3$--$7$   & $10$--$25$ \\
    \bottomrule
  \end{tabular}
\end{table}

\subsection{Closed-form Bekker--Wong terramechanics}
\label{sec:terramechanics}
We implement a rigid-wheel Bekker--Wong kernel that integrates the normal and shear stresses around the wheel--soil contact patch, following the rigid-wheel formulation of Wong and Reece \cite{wongPredictionRigidWheel1967} augmented by a simple engaged-grouser shear correction motivated by single-wheel grouser experiments \cite{iizukaEffectTractiveGiven2011} and made determinate by a vertical force balance that fixes the sinkage. Each wheel is treated as a rigid disk of radius $r$ and width $b$ that sinks to a depth $z_0$ into regolith. The contact patch runs from an entry angle $\theta_1$ (where the rim first meets the undisturbed surface) to an exit angle $\theta_2$, taken as $\theta_2 = 0$ following the convention that elastic rebound behind the wheel contributes no net stress.

\paragraph{Stress distribution.}
With $\theta$ measured from the downward vertical, the soil intrusion depth is $z(\theta) = r(\cos\theta - \cos\theta_1)$, with maximum sinkage $z_0 = r(1-\cos\theta_1)$. Substituting into Bekker's law~\eqref{eq:bekker} gives the radial normal stress $\sigma(\theta)$, which peaks at a transition angle $\theta_m = (c_1 + c_2\,|s|)\,\theta_1$, with $c_1=0.4$, $c_2=0.2$ \cite{wongPredictionRigidWheel1967}. This splits the contact patch into a front region ($\theta_m \le \theta \le \theta_1$) and a rear region ($0 \le \theta < \theta_m$),
\begin{equation}
  \sigma(\theta) =
  \begin{cases}
    k_{\mathrm{eff}}\,r^{n}\,(\cos\theta - \cos\theta_1)^{n}, & \theta_m \le \theta \le \theta_1,\\[4pt]
    k_{\mathrm{eff}}\,r^{n}\,(\cos\theta^\star - \cos\theta_1)^{n}, & 0 \le \theta < \theta_m,
  \end{cases}
  \label{eq:sigma}
\end{equation}
with $k_{\mathrm{eff}} = k_c/b + k_\phi$. The rear region reuses the front-region shape via the linear remap $\theta^\star(\theta) = \theta_1 - (\theta/\theta_m)(\theta_1 - \theta_m)$, so $\sigma$ is continuous at $\theta_m$ and vanishes at the exit. The shear displacement of a soil element first engaged at $\theta_1$ is $j(\theta) = r[(\theta_1-\theta) - (1-s)(\sin\theta_1-\sin\theta)]$, with slip ratio $s$ from~\eqref{eq:slip}. Equation~\eqref{eq:janosi} gave the Janosi--Hanamoto shear law in its generic point form; the version used by the kernel resolves it around the contact arc and extends it in two ways,
\begin{equation}
  \tau(\theta) = \gamma_g\,\bigl(c + \sigma(\theta)\tan\phi\bigr)
                 \Bigl(1 - e^{-|j(\theta)|/K}\Bigr)\,\operatorname{sgn} j(\theta).
  \label{eq:tau}
\end{equation}
First, the normal stress $\sigma(\theta)$ and shear displacement $j(\theta)$ are now functions of contact angle rather than scalars, so $\tau$ varies along the patch and enters the wheel-force integrals below. Second, motivated by single-wheel grouser experiments \cite{iizukaEffectTractiveGiven2011}, the kernel adds the engaged-grouser prefactor $\gamma_g$ and the $\operatorname{sgn}(j(\theta))$ factor that keeps the shear stress sign correct across the full slip range. This grouser-augmented form, rather than the generic law of~\eqref{eq:janosi}, is what the kernel integrates to obtain the wheel forces.

\paragraph{Grouser lift.}
Because micro-rover wheels are typically grousered, we apply a bounded empirical arc-density correction $\gamma_g$ to the shear thrust in the kernel,
\begin{equation}
  \gamma_g = 1 + \min\!\left(\frac{N_g\,h_g}{2\pi r},\; \gamma_{g,\max}\right),
  \label{eq:grouser}
\end{equation}
with grouser count $N_g$, grouser height $h_g$, and a cap $\gamma_{g,\max}=0.6$ that keeps the correction physical once adjacent shear planes would interfere; $\gamma_g=1$ recovers the smooth-wheel kernel and is $\theta_1$-independent, so it factors cleanly out of the contact integrals. 

\paragraph{Force balance and outputs.}
Resolving the normal and shear stresses along the contact arc (arc element $r\,\mathrm{d}\theta$, width $b$) and integrating over $\theta\in[0,\theta_1]$ yields the vertical load $F_z$, the net drawbar pull $\mathrm{DP}$, and the driving torque $T = b\,r^{2}\!\int_0^{\theta_1}\tau\,\mathrm{d}\theta$. For a given per-wheel vertical load $F_z^\star$ representing the rover weight under lunar gravity shared across the wheels, the entry angle $\theta_1$ is fixed by the force-balance condition $F_z(\theta_1)=F_z^\star$, which we solve over $\theta_1\in(0,\pi/2)$ with Brent's method. Sinkage is therefore recovered as $z_0 = r(1-\cos\theta_1)$ and is an output of the kernel, not an input to it.

The kernel takes wheel geometry, soil parameters, the per-wheel vertical load $F_z^\star$, and the slip ratio $s$, and returns drawbar pull, driving torque, sinkage, and the entry angle. Slip is an independent input at this level, as is conventional for Bekker--Wong models, but it is not left free at the mission level: the traverse simulation of Section~\ref{sec:traverse} solves for the slip that develops the drawbar pull the scenario slope demands, so both $s$ and $z_0$ are determined by the design and the scenario rather than assumed. The resulting per-wheel drawbar pull and torque drive the slope capability and the traverse energy budget (Section~\ref{sec:traverse}).

\paragraph{Soil parameters and runtime.}
The soil parameters $(n, k_c, k_\phi, c, \phi, K)$ are assigned per terrain class from a simulant table compiled from public data on simulant parameters \cite{kanamoriPropertiesLunarSoil2012,zengGeotechnicalPropertiesJSC1A2010,oravecDesignCharacterizationGRC12010,heGeotechnicalPropertiesGRC32013,heiken1991,dingExperimentalStudyAnalysis2011,wangcheng-canExperimentalStudyLunar2016,hurrellTractionPerformanceEvaluation2025,ozakiGranularFlowExperiment2023}, with the shear-deformation modulus defaulting to $K=0.018$\,m \cite{heiken1991}. The contact-arc integrals are evaluated on a 100-point trapezoidal grid with error well below the $\pm20$--$30\,\%$ Bekker--Wong model-form band \cite{wuExperimentalStudyAnalysis2026, agarwalModelingInteractionRigid2019,ishigamiTerramechanicsbasedModelSteering2007} while costing a median $0.50$\,ms per wheel evaluation on the reference machine of Section~\ref{sec:runtime}, making it fast enough for the kernel to run directly inside the optimizer's fitness loop (Section~\ref{sec:moo}). As argued in Section~\ref{sec:background-theory}, a physics-based numerical contact model would be more appropriate for detailed wheel design but would move the contact model outside the optimizer fitness loop. 

\subsection{Mass model}
\label{sec:mass-model}
The mass model is bottom-up: each subsystem is sized from a directly traded design variable and a fixed specific-mass constant or sizing fraction. The rollup convention follows Space Mission Analysis and Design (SMAD) and AIAA~S-120A mass-properties practice \cite{smad2011,aiaa_s120a}: harness and thermal-control mass are applied as fractions of the summed subsystem mass, the dry-mass growth allowance is applied to the dry bus, and the scientific payload is added afterward because it is specified by the mission scenario directly. For a design $x$, the dry bus mass is assembled as
\begin{align}
  m_\mathrm{sub} &=
      m_c + m_w + m_m + m_s + m_b + m_a + m_{\mathrm{arch}}, \label{eq:mass-subsystems}\\
  m_\mathrm{dry} &=
      m_\mathrm{sub}
      + f_h m_\mathrm{sub}
      + f_t\!\left(m_\mathrm{sub}+f_h m_\mathrm{sub}\right), \label{eq:mass-dry}\\
  m_\mathrm{total} &=
      m_\mathrm{dry}
      + f_g m_\mathrm{dry}
      + m_\mathrm{payload}. \label{eq:mass-total}
\end{align}
$m_c$ is the chassis structural mass, $m_w$ the wheel and grouser mass, $m_m$ the motor/gearbox mass, $m_s$ the solar-array mass, $m_b$ the battery mass, $m_a$ the avionics mass, and $m_{\mathrm{arch}}$ the rocker-bogie suspension/linkage mass (zero for a rigid four-wheel layout). The fractions $f_h$, $f_t$, and
$f_g$ are the harness, thermal-control, and dry-mass growth allowances, respectively. The scientific payload mass is added after the dry-mass growth allowance because it is specified by the mission scenario rather than being included in the input design vector as a variable to trade.

The subsystem terms are defined as:
\begin{align}
  m_w &= N_w\left[
      \rho_A (2\pi r b)
      + N_g\,b\,h_g\,t_g\,\rho_g
      \right], \label{eq:mass-wheels}\\
  m_m &= N_w\left(m_{m,0} + k_\tau T_{\mathrm{hub}}^{\mathrm{peak}}\right),
      \label{eq:mass-motors}\\
  m_s &= \rho_s A_s,\qquad
  m_b = \frac{C_b}{e_b},\qquad
  m_a = m_{a,0} + k_a P_a . \label{eq:mass-power-avionics}
\end{align}
where $N_w$ is the drive-wheel count; $\rho_A$ is the wheel-side areal mass density (rim, hub, spokes, and fasteners) and $r$, $b$ the wheel radius and width. $N_g$, $h_g$, and $t_g$ are the grouser count, height, and plate thickness, and $\rho_g$ is the grouser material density. In the motor term, $m_{m,0}$ is the per-wheel motor/gearbox base mass, $k_\tau$ the mass per unit of peak output torque, and $T_{\mathrm{hub}}^{\mathrm{peak}}$ the design's peak per-wheel hub torque. For the remaining subsystems, $\rho_s$ is the solar-array areal mass density and $A_s$ its planform area; $C_b$ and $e_b$ are the battery
energy capacity and pack-level specific energy; and $m_{a,0}$, $k_a$, and $P_a$ are the avionics base mass, mass-per-watt allowance, and continuous avionics power. That avionics term is a lumped C\&DH box: $m_{a,0}$ is a floor for enclosure, backplane, and one CPU card, and $k_a P_a$ adds heat-sink and enclosure mass in proportion to continuous draw. Science instruments sit in $m_{\mathrm{payload}}$; solar, battery, motors, harness, and thermal-control hardware are separate line items. Other bus electronics (GNC, comms, navigation cameras) enter only insofar as their continuous draw is included in $P_a$.

The wheel term scales rim structure with cylindrical side area and adds each grouser as a thin aluminium plate; the motor term sizes a motor/gearbox package from the design's peak hub torque; the remaining terms use solar-array areal mass, battery pack specific energy, and an avionics mass floor plus an allowance for the heat sink and enclosure. For a rocker-bogie layout, $m_{\mathrm{arch}} = 0.5\,\mathrm{kg} + 0.08\,m_c$ is charged before harness and growth; those two coefficients are an engineering assumption for the micro-rover class, not a fit to published linkage masses. All coefficients are exposed as model parameters so that sensitivity studies can perturb the mass model without changing the evaluator's structure.

\subsection{Power and thermal}
\label{sec:power-thermal}
Solar input is computed with a closed-form lunar solar geometry model \cite{duffiebeckman2013}. For latitude $\lambda$, solar declination $\delta$, and local hour angle $H$, the sun elevation is
\begin{equation}
  \sin e = \sin\lambda\sin\delta + \cos\lambda\cos\delta\cos H .
  \label{eq:sun-elevation}
\end{equation}
For a flat plate of area $A_s$ tilted by $\beta$ toward azimuth $\psi$, the incidence factor is
\begin{equation}
  \cos i =
  \sin e \cos\beta + \cos e \sin\beta \cos(\alpha_\odot-\psi),
  \label{eq:solar-incidence}
\end{equation}
where $\alpha_\odot$ is the solar azimuth. The instantaneous electrical power is
\begin{equation}
  P_\mathrm{solar} =
  S_0 A_s \eta_s d_s \max(0,\cos i),
  \label{eq:solar-power}
\end{equation}
with $S_0=1361\,\mathrm{W\,m^{-2}}$ \cite{kopplean2011}, solar conversion efficiency $\eta_s$, and dust/degradation factor $d_s$. The default array is horizontal; high-latitude runs can pass a fixed tilt, typically $\min(80^\circ,|\lambda|)$, to represent a deployable panel aligned with the low-elevation polar sun. The model captures the latitude-driven sun-angle penalty but does not account for terrain horizon masking or site shadowing, so high-latitude results isolate the sun-angle component of the polar energy problem (Section~\ref{sec:limitations}).

The battery model is a coulomb-counting state of charge (SOC) update with a depth of discharge floor \cite{smad2011}. For net bus power into the battery $P_\mathrm{net}$ and step duration $\Delta t$,
\begin{equation}
  \Delta E =
  \begin{cases}
    \eta_c P_\mathrm{net}\Delta t, & P_\mathrm{net}\ge 0,\\
    P_\mathrm{net}\Delta t/\eta_d, & P_\mathrm{net}<0,
  \end{cases}
  \label{eq:battery-update}
\end{equation}
where $\eta_c$ and $\eta_d$ are the charge and discharge efficiencies (both $0.95$ in the default model), and the SOC is clamped between the minimum allowed SOC (15\% in the default model) and full charge. Surplus generation at full charge is discarded: the array keeps producing, the battery does not accept the extra energy, and there is no dump load or extra mobility draw. The rover does not shut down at the floor: Section~\ref{sec:traverse} cuts mobility duty to whatever solar remains after the avionics and payload load, and the run continues to the end of the mission.

Thermal survival is represented by a single-node steady-state enclosure model \cite{gilmore2002}. The enclosure is one thermal node: absorbed solar, avionics and payload dissipation, and any radioisotope heater unit (RHU) power enter as a single $Q_\mathrm{in}$, and heat leaves only by radiation to $T_{\mathrm{sink}}$. There is no conduction between components, no conduction to the ground, and no convection. For radiating area $A_r$, emissivity $\epsilon$, and effective radiative sink temperature $T_\mathrm{sink}$, the equilibrium node temperature is
\begin{equation}
  T =
  \left(
    T_\mathrm{sink}^4 +
    \frac{Q_\mathrm{in}}{\epsilon\sigma_\mathrm{SB}A_r}
  \right)^{1/4}.
  \label{eq:thermal-node}
\end{equation}
The evaluator checks a hot case of peak solar absorption plus avionics and payload power against $T_{\mathrm{sink}}=250$\,K and a cold case of lunar night with hibernation load and any radioisotope heater unit (RHU) power against $T_{\mathrm{sink}}=100$\,K. Those sinks are effective environment temperatures for a small enclosure viewing a large mix of lunar surface and sky. Incoming infrared from that environment is already in the $T_{\mathrm{sink}}^4$ term, so a separate sink absorptivity does not appear. The night value follows the regolith surface temperature directly, while the peak-sun value is a view-factor weighting between the hot subsolar surface below the rover and the cold sky above, taking surface extremes of $\approx390$\,K subsolar and $\approx100$\,K at night \cite{heiken1991}. Section~\ref{sec:flown-consistency} checks the resulting flag against the published lunar-night outcomes of two flown rovers. The result is reported as a survival flag rather than used as an optimizer objective or constraint. With the heater-free default enclosure of the class-generic scenarios every design fails the cold case, consistent with micro-rovers carrying no radioisotope heaters not surviving lunar night. It discriminates only when a rover's own heater configuration is supplied, which is how Section~\ref{sec:flown-consistency} uses it, so the reported fronts are not filtered on thermal survival.

\subsection{Traverse simulation}
\label{sec:traverse}
The traverse simulation ties the other sub-models together in time. From the total mass of Eq.~\eqref{eq:mass-total} and the scenario slope $\theta$, it computes the per-wheel normal load and required drawbar pull, dividing both equally among the $N_w$ wheels,
\begin{equation}
  F_z^\star = \frac{m_\mathrm{total} g \cos\theta}{N_w},
  \qquad
  \mathrm{DP}^\star = \frac{m_\mathrm{total} g \sin\theta}{N_w},
  \label{eq:required-dp}
\end{equation}
and solves the slip balance
\begin{equation}
  \mathrm{DP}(s; F_z^\star, r, b, h_g, N_g, \mathrm{soil})
  - \mathrm{DP}^\star = 0
  \label{eq:slip-balance}
\end{equation}
over a bounded slip interval. The rover is considered to be stalled if no slip can develop the required pull or if the solution's torque exceeds the design's peak hub torque $T_{\mathrm{hub}}^{\mathrm{peak}}$. Slope therefore enters the kernel only through these two terms, which reduce the normal load by $\cos\theta$ and set the pull each wheel must develop through $\sin\theta$; the kernel itself is written for a single wheel and is not slope-aware.

Dividing the load equally assumes a symmetric static weight distribution. Longitudinal load transfer, which shifts weight onto the downhill wheels in proportion to center-of-mass height, is not modeled, nor is a center-of-mass offset within the chassis. Because drawbar pull is nearly linear in normal load over this range, redistributing weight at fixed total mass changes the total available pull by under $2\,\%$ across the reported fronts for center-of-mass-height-to-wheelbase ratios up to $0.4$ at a $25^\circ$ slope, an order of magnitude inside the kernel's own model-form band. The omission matters more for tip-over stability, which is outside the current constraint set (Section~\ref{sec:limitations}).

Cruise speed is derived for non-stalled designs. The drivetrain model first computes the energy-balance speed at which average solar power supports avionics, payload, and duty-cycled mobility,
\begin{equation}
  v_\mathrm{eb} =
  \frac{
    (\bar{P}_\mathrm{solar}-P_a-P_\mathrm{payload})
    r(1-s)\eta_m
  }{
    \delta_\mathrm{eff} N_w T
  },
  \label{eq:energy-balance-speed}
\end{equation}
where $\bar{P}_\mathrm{solar}$ is the mission-average solar power, $P_\mathrm{payload}$ is the continuous payload draw, $T$ is the per-wheel driving torque from the terramechanics solve, $\eta_m$ is drivetrain efficiency, and $\delta_\mathrm{eff}=\mathrm{clamp}(\delta_\mathrm{ops},0,1)$ with $\delta_\mathrm{ops}$ the operational duty cycle. Cruise speed is $v_{\mathrm{eb}}$, or zero if the rover is stalled.

The simulation then advances with a one-hour time step over the duration of the mission scenario. At each step the solar model supplies $P_\mathrm{solar}(t)$, the load sums avionics, payload, and duty-cycled mobility power, Eq.~\eqref{eq:battery-update} advances SOC, and position increments by $v_{\mathrm{eb}}\delta_\mathrm{eff}\Delta t$ unless stalled. If the battery hits its SOC floor while load exceeds solar input, mobility duty is throttled to
the solar-supported fraction rather than terminating the run, so that the traverse simulation always reaches the end of the mission and every design returns a range the optimizer can rank. The reported range is thus an energy-feasible mission range under the scenario duty cycle. Every run uses a fixed illumination timing: zero solar declination ($\delta=0$) and a mission start pinned to local sunrise, so the solar profile $P_\mathrm{solar}(t)$ is determined by latitude, mission duration, and panel orientation alone. 

Because the current scenarios use fixed slope and soil as a conservative assumption, the expensive slip-balance solve is loop-invariant and is lifted out of the time loop, leaving the per-step update dominated by solar geometry and battery accounting.

\paragraph{Computational cost.}
\label{sec:runtime}
All timings reported in this paper were measured single-threaded on one core of an Apple M2 Pro (macOS~15, Python~3.12.13, NumPy~2.4.3, pymoo~0.6.1.6). Cost is characterized over $400$ designs drawn uniformly from the search bounds of Table~\ref{tab:design-ranges} and decoded through the same path the optimizer uses, so the measured distribution is the one NSGA-II pays for, including infeasible designs. A complete mission evaluation costs a median $10.3$\,ms on the $14$-day equatorial mare scenario ($9.5$--$11.6$\,ms across the four scenario families of Table~\ref{tab:design-ranges}, $95$th percentile $12.8$\,ms), or $86$--$105$ mission evaluations per second on a single core.

Because the slip-balance solve is lifted out of the time loop, this cost is dominated by a term independent of mission length. Fitting per-mission cost against simulated duration over $1$--$365$ Earth days gives $9.1$\,ms of fixed cost plus $3.6$\,$\mu$s per simulated hour: a one-day mission costs $9.2$\,ms and a one-year mission $41$\,ms, and only $1.2$ of the $10.3$\,ms of a $14$-day evaluation is spent in the time loop. On the same $400$-design sample the slope-capability solve is $3.3$\,ms and the traverse call is $6.8$\,ms, of which that $1.2$\,ms is the time-loop solar and battery update; the mass and thermal models together are under $0.01$\,ms. Runtime is therefore set by the Bekker--Wong solves, not by mission duration or by the other subsystem models.

Counting kernel invocations on the same sample shows how few of those solves an evaluation needs. A mission evaluation performs a median of $19$ wheel--soil solves---$7$ in the slope-capability bisection and $12$ in the traverse slip solve---rather than one per time step, which for a $14$-day mission at a one-hour step would be $336$. Hoisting the solve out of the time loop therefore removes roughly a factor of $18$. It does not relax the need for a closed-form kernel, because the count that matters is per design rather than per time step: one Pareto front of $3\,000$ evaluations still costs about $57\,000$ wheel--soil solves. At the hours-per-run cost of the numerical contact models of Section~\ref{sec:background-theory}, substituting one of them would exceed the present budget by many orders of magnitude even at this reduced call count. End to end, one NSGA-II front of $50$ individuals over $60$ generations ($3\,000$ mission evaluations) completes in $37$\,s on a single core, and the four-scenario study of Section~\ref{sec:discussion} in under three minutes---fast enough to map Pareto fronts directly inside the optimizer loop. The benchmark script and its recorded output are released with the tool so these figures can be reproduced on other hardware.

\section{Optimizer}
\label{sec:optimizer}
The analytical evaluator is fast enough to serve directly as the NSGA-II fitness function, and it is the ground truth for every Pareto front and scientific result in this paper.

\subsection{Multi-objective optimization}
\label{sec:moo}
The tradespace is searched with NSGA-II \cite{debFastElitistMultiobjective2002} to identify the Pareto front of non-dominated designs, the set of achievable compromises among the objectives. The $11$-dimensional design vector is optimized within a continuous box whose bounds mirror the schema of Table~\ref{tab:design-ranges}. We optimize three competing objectives which form the core mission-level trade in the micro-rover regime: maximize range, minimize total mass, and maximize slope capability. The mission scenario is held fixed within a run, so each scenario yields its own front; the optimizer searches designs, not missions.

\paragraph{Implementation.}
The optimizer is the NSGA-II implementation in pymoo~$0.6.1.6$ \cite{blankPymooMultiObjectiveOptimization2020}, invoked with its default operators; Table~\ref{tab:nsga2-settings} states them so that runs can be reproduced. Objectives are converted to minimization form internally by negating the two maximized targets. A single inequality constraint imposes a $0.1$\,km range floor, expressed as a violation magnitude so that the algorithm can rank infeasible individuals by how badly they miss rather than treating all of them alike. Selection is a binary tournament that prefers the smaller constraint violation when either competitor is infeasible and otherwise the better Pareto rank, breaking ties by crowding distance, the normalized spacing to a design's neighbors on its front, which favors designs in sparsely populated regions and so keeps the front spread out.

Nine of the eleven design variables are continuous. The other two are not: grouser count is an integer, and mobility architecture is categorical. Both are handled by decoding rather than by a mixed-variable algorithm. NSGA-II searches a real-valued gene for each, and that gene is converted to a valid discrete value before the design is evaluated: the grouser gene is rounded to the nearest integer and clipped to $[0,24]$, and the architecture gene is thresholded at the midpoint of its range to select either the rigid four-wheel or the rocker-bogie six-wheel layout, which in turn fixes the wheel count. Because decoding precedes evaluation, every objective value the algorithm sees corresponds to a rover that could physically be built, and no fractional wheel or grouser count can reach the reported fronts.
\begin{table}[htbp]
  \centering
  \small
  \caption{NSGA-II implementation. Operator settings are the pymoo~$0.6.1.6$ defaults; SBX and polynomial-mutation distribution indices follow the convention of \cite{debFastElitistMultiobjective2002}.}
  \label{tab:nsga2-settings}
  \begin{tabular}{@{}ll@{}}
    \toprule
    Setting & Value \\
    \midrule
    Library & pymoo $0.6.1.6$ \cite{blankPymooMultiObjectiveOptimization2020} \\
    Encoding & $11$ real-valued genes in a box; $2$ decoded to discrete \\
    Population size & $50$ \\
    Offspring per generation & $50$ \\
    Generations & $60$ (canonical); $30$/$60$/$90$ for the budget check \\
    Initial sampling & Uniform random within bounds \\
    Selection & Binary tournament (pressure $2$) \\
    Crossover & Simulated binary (SBX), $p_c = 0.9$, $\eta_c = 15$ \\
    Mutation & Polynomial, $p_m = 0.9$ per individual, \\
      & $1/n_\mathrm{var} = 0.09$ per gene, $\eta_m = 20$ \\
    Survival & Non-dominated rank, then crowding distance \\
    Duplicate elimination & Enabled \\
    Objectives & $3$ (max range, min mass, max slope capability) \\
    Constraints & $1$ (range $\ge 0.1$\,km) \\
    Failed evaluations & Assigned a strongly infeasible sentinel \\
    Seeds & $2$ per scenario (generation-budget check) \\
    \bottomrule
  \end{tabular}
\end{table}

The canonical fronts use a population of $50$ evolved over $60$ generations for each scenario. We repeated each run at $30$, $60$, and $90$ generations for two independent seeds to check that this budget is enough. Normalized hypervolume (the volume of objective space a front dominates, after scaling range, mass, and slope capability to a common box) is used here only as a convergence check. It reaches $99.3$--$99.6\,\%$ of the $90$-generation value by generation $60$ (Table~\ref{tab:optimizer-robustness}), so extra generations add little. 

\begin{table}[htbp]
  \centering
  \small
  \caption{NSGA-II generation-budget check. Hypervolume values are means across two independent seeds.}
  \label{tab:optimizer-robustness}
  \begin{tabular}{lcccc}
    \toprule
    Scenario & HV$_{30}$ & HV$_{60}$ & HV$_{90}$ & HV$_{60}$/HV$_{90}$ \\
    \midrule
    Crater-rim survey & $0.782$ & $0.794$ & $0.799$ & $0.993$ \\
    Equatorial mare traverse & $0.782$ & $0.796$ & $0.801$ & $0.993$ \\
    Highland slope capability & $0.469$ & $0.483$ & $0.487$ & $0.993$ \\
    Polar prospecting & $0.766$ & $0.778$ & $0.781$ & $0.996$ \\
    \bottomrule
  \end{tabular}
\end{table}

\section{Validation methodology and results}
\label{sec:validation}
We assess the tool as a multi-layer evidence stack rather than claiming a single end-to-end validation, because its components carry different evidentiary weight (Table~\ref{tab:validation-ledger}). Empirical support for the physics-based evaluator rests on two component-level, two-sided comparisons against external ground truth: the terramechanics kernel against measured single-wheel data (Section~\ref{sec:terra-validation}) and the bottom-up mass model against published rover masses (Section~\ref{sec:mass-validation}). Here ``two-sided'' denotes a comparison that can fail by over- or under-prediction, in contrast to a one-sided feasibility check that is trivially satisfied in one direction. The remaining layers---flown-rover consistency (Section~\ref{sec:flown-consistency}) and the system-level rediscovery check (Section~\ref{sec:rediscovery})---provide supporting credibility evidence for particular sub-models or for the integrated tool.

\begin{table}[htbp]
  \centering
  \small
  \setlength{\tabcolsep}{4pt}
  \caption{Validation stack for the evaluator.}
  \label{tab:validation-ledger}
  \begin{tabular}{@{}>{\raggedright\arraybackslash}p{0.20\linewidth}>{\raggedright\arraybackslash}p{0.42\linewidth}>{\raggedright\arraybackslash}p{0.28\linewidth}@{}}
    \toprule
    Validation Layer & Comparison & Evidentiary role \\
    \midrule
    Terramechanics (\S\ref{sec:terra-validation}) &
      Closed-form rigid-wheel kernel vs. measured single-wheel drawbar pull and sinkage, three independent sources &
      Empirical, two-sided \\[2pt]
    Mass model (\S\ref{sec:mass-validation}) &
      Bottom-up total mass vs.\ published total mass, five in-class rovers
      (median $|\text{err}|$ $10.5\,\%$) &
      Empirical, two-sided \\[2pt]
    Flown rovers (\S\ref{sec:flown-consistency}) &
      Fixed-parameter peak solar, thermal survival flag, and range envelope vs.\
      Pragyan and Yutu-2 &
      Real-world consistency (one-sided) \\[2pt]
    Rediscovery (\S\ref{sec:rediscovery}) &
      Optimizer Pareto fronts vs.\ real rover designs ($n=5$) &
      System-level credibility \\
    \bottomrule
  \end{tabular}
\end{table}

\subsection{Terramechanics validation against measured single-wheel data}
\label{sec:terra-validation}
Among the sub-models we benchmark, the wheel--soil kernel has the largest model-form error (median drawbar-pull error $24$--$27\,\%$ on the in-regime single-wheel sets, versus $10.5\,\%$ for mass and $+5\,\%$ for Pragyan peak solar). We therefore establish its credibility against measured single-wheel data. Drawbar-pull and sinkage-versus-slip from the kernel were evaluated against measurements from three independent published testbeds to record per-point error. 

\begin{figure}[htbp]
  \centering
  \includegraphics[width=\linewidth]{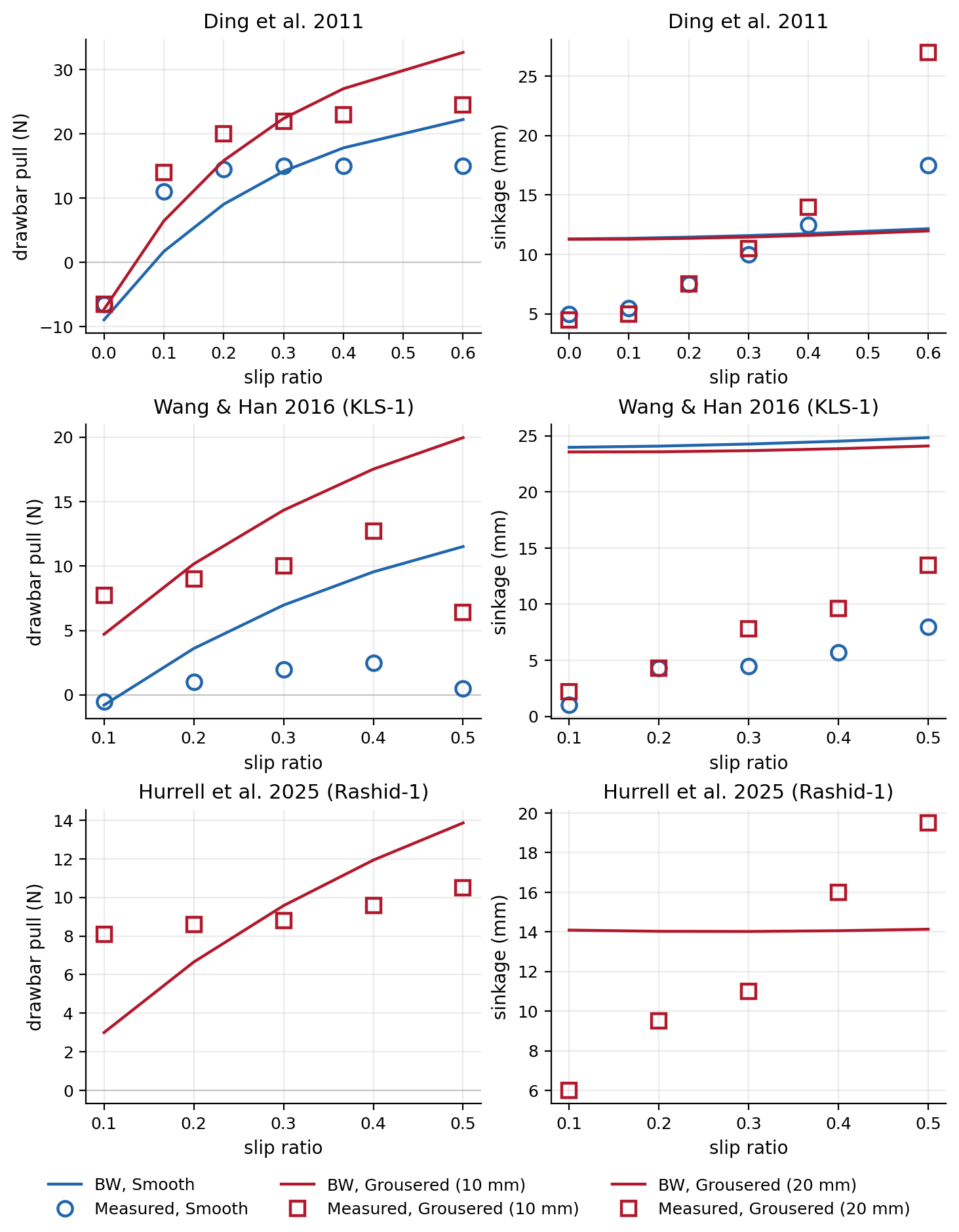}
  \caption{Terramechanics validation against measured single-wheel data:
           drawbar pull and sinkage vs.\ slip for Ding et al.\ \cite{dingExperimentalStudyAnalysis2011},
           Wang \& Han \cite{wangcheng-canExperimentalStudyLunar2016} (KLS-1), and Hurrell et al.\
           \cite{hurrellTractionPerformanceEvaluation2025} (Rashid-1), smooth and grousered.}
  \label{fig:terramechanics}
\end{figure}

The three sources study complementary regimes (Fig.~\ref{fig:terramechanics}). Ding et al.\ \cite{dingExperimentalStudyAnalysis2011} tested smooth and grousered rigid wheels ($r=157$\,mm, $80$\,N) on a simulant whose full Bekker parameter set is reported. The kernel matches drawbar pull to a median $27\,\%$ and reproduces the grouser traction gain, but under-predicts slip-dependent sinkage (median $52\,\%$), a known limitation of the rigid-wheel formulation. Hurrell et al.\ \cite{hurrellTractionPerformanceEvaluation2025} characterized the in-scope Rashid-1 micro-rover wheel ($r=100$\,mm, $14$ grousers, $24.5$\,N, FJS-1 simulant), the most relevant application case, and the kernel matches drawbar pull to $24\,\%$ and sinkage to $28\,\%$. Wang \& Han \cite{wangcheng-canExperimentalStudyLunar2016, limDevelopmentNewPressureSinkage2021} ($r=85$\,mm, $59$\,N, KLS-1) is a stress case at the edge of the kernel's regime. It pairs a small, lightly loaded wheel with a firm, dense simulant (KLS-1 compacted to $\sim$60\,\% relative density) so the wheel barely sinks ($1$--$14$\,mm). The rigid-wheel kernel structurally cannot match this: because sinkage is solved implicitly from vertical force balance, the contact patch required to carry $59$\,N on an $85$\,mm wheel drives the solver to $\sim$24\,mm almost independently of the soil moduli, so the kernel over-predicts both drawbar pull and sinkage (median $135\,\%$ and $385\,\%$).

In summary, drawbar pull predictions from our kernel on the two in-regime sources (Ding and Hurrell) are consistent with the range of per-point errors typically reported for Bekker-Wong models against single wheel data of $\pm20$--$30\,\%$ \cite{ishigamiTerramechanicsbasedModelSteering2007,wuExperimentalStudyAnalysis2026,agarwalModelingInteractionRigid2019}, while the KLS-1 case bounds the kernel's regime of validity. The KLS-1 regime---a small, lightly loaded wheel on a firm, densely compacted simulant (relative density $\sim$60\,\%) in which the wheel barely sinks---is firmer and higher-bearing-strength than the nominal-to-loose regolith of the four mission scenarios, so it lies outside the soft-regolith sinkage regime that governs the sub-$50$\,kg traverse missions targeted here; Ding's planetary simulant and Hurrell's loose FJS-1 are therefore the relevant benchmarks for those conclusions. We propagate this drawbar-pull error into the optimizer's conclusions in Section~\ref{sec:terra-sensitivity}.

\subsection{Mass-model validation against published masses}
\label{sec:mass-validation}
The bottom-up mass model (Section~\ref{sec:mass-model}) is compared against published rover masses with each rover's scientific payload entered as a line item so the comparison is on the same bus basis. The check is on total mass only. Published sources for this class rarely report a consistent subsystem breakdown, so a matching total does not confirm that the allocation among structure, mobility, power, and avionics is correct. The comparison is a conceptual-design sanity check on the rollup, not a validation of the internal mass budget. The comparison set spans both lunar and Mars rovers, since the model's specific-mass coefficients size structure, drivetrain, avionics, and power subsystems and are effectively gravity-independent. Median absolute error is reported over the five in-class ($5$--$50$\,kg) rovers; out-of-regime cases are shown but excluded from the figure metric.

Across the in-class rovers the model predicts total mass to a median absolute error of $10.5\,\%$ (mean $10.0\,\%$), inside the typical margins during the conceptual design phase (Table~\ref{tab:mass-validation}). Individual errors span $0.7$--$19.9\,\%$ and do not separate by mobility architecture: the three four-wheel cases are Tenacious $+19.9\,\%$, CADRE $-5.7\,\%$, and Rashid $+10.5\,\%$, and the two six-wheel cases are Sojourner $-0.7\,\%$ and Pragyan $-13.3\,\%$. What the errors track is mass. The lightest in-class rover is over-predicted and the heaviest is under-predicted, and the same pattern continues into the out-of-regime rows. The two ends have different causes. At the light end the mass-independent terms---the per-wheel motor and avionics base masses and the fixed rocker-bogie linkage---do not shrink with the vehicle; grossed up by the harness, thermal, and growth fractions they reach a quarter of published mass at Tenacious's $5$\,kg. At the heavy end the model carries no line item for hardware that appears only above the micro class: steering actuators beyond the drive motors, deployable-array and antenna-mast mechanisms, a telecom subsystem, and the radioisotope heaters and heat-pipe hardware lunar-night survival requires, none of it captured by a flat thermal fraction. That is why Yutu-2 and MARSOKHOD are under-predicted by $8$--$13\,\%$.

\begin{table}[htbp]
  \centering
  \small
  \caption{Mass-model validation: bottom-up predicted vs.\ published full-up
           total mass. Out-of-regime rovers (italic, $>50$\,kg) are reported
           but excluded from the aggregate error statistics. MARSOKHOD's
           published mass is the low end of the $70$--$75$\,kg range in
           the source.}
  \label{tab:mass-validation}
  \begin{tabular}{@{}lrrrl@{}}
    \toprule
    Rover & Published & Predicted & Error & Source \\
          & (kg)      & (kg)      & (\%)  & \\
    \midrule
    Tenacious            &  5.0 &   6.0 & $+19.9$ & \cite{IspaceEUROPEAnnouncesCompletion} \\
    CADRE-unit           &  8.6 &   8.1 & $-5.7$ & \cite{nasaCadreFactSheet} \\
    Rashid               & 10.0 &  11.0 & $+10.5$ & \cite{alzaabiOperationalConceptsRehearsal2024} \\
    Sojourner            & 11.5 &  11.4 & $-0.7$ & \cite{wilcoxSojournerMarsLessons1998} \\
    Pragyan              & 26.0 &  22.5 & $-13.3$ & \cite{Chandrayaan3EoPortal} \\
    \midrule
    \emph{Yutu-2} (\textgreater50\,kg)       & \emph{135.0} & \emph{117.6} & \emph{$-12.9$} & \cite{tangPhysicalMechanicalCharacteristics2020} \\
    \emph{MARSOKHOD} (\textgreater50\,kg)     & \emph{70.0} & \emph{64.2}  & \emph{$-8.2$} & \cite{kemurdjianSmallMarsokhodConfiguration1992} \\
    \midrule
    \multicolumn{5}{@{}l@{}}{\emph{In-class median $|$error$|$ = 10.5\,\%; mean = 10.0\,\%.}} \\
    \bottomrule
  \end{tabular}
\end{table}

\subsection{Flown-rover consistency}
\label{sec:flown-consistency}
For the two flown rovers (Pragyan, Yutu-2) we run two additional one-sided consistency checks. First, the thermal-survival flag: the single-node hot/cold model reproduces both published outcomes, correctly predicting that Pragyan (no radioisotope heater units) does not survive the lunar night while the RHU-equipped Yutu-2 does. Second, the range envelope: because the evaluator assumes constant-speed driving at the duty-cycle floor and omits the long science and thermal-wait windows of real operations, its range is a capability upper bound rather than an operational estimate; the predicted envelope exceeds the published operational distance for both rovers ($500$ vs.\ $101$\,m for Pragyan, $200$ vs.\ $25$\,m for Yutu-2). We treat both as corroboration of the thermal and traverse sub-models rather than two-sided accuracy validation.

Predicted peak solar power is also compared against the reported values for the two flown rovers. A common beginning-of-life (BOL) panel parameter set is applied to every rover, with only published array area and landing latitude varied by rover. The net system efficiency is $\eta_\mathrm{sys}=0.30\times0.90\times0.92\times0.90\approx0.224$, built from a cell efficiency of $30\,\%$, typical of space-grade solar cells at beginning of life, a $90\,\%$ active-area packing factor and $92\,\%$ electrical efficiency (peak-power tracking, harness, and blocking diodes) following standard power-system practice \cite{patelSpacecraftPowerSystems2017,smad2011}, and a $10\,\%$ lunar-noon temperature derate obtained from a gallium-arsenide power coefficient of $-0.06\,\%$\,K$^{-1}$ and a cell running $\approx90$\,K above the air-mass-zero reference temperature. We also assume a nominal $2\,\%$ optical loss on a fresh array. Cell efficiency is the only term the prediction is materially sensitive to, and it is swept below.

\begin{figure}[htbp]
  \centering
  \includegraphics[width=\linewidth]{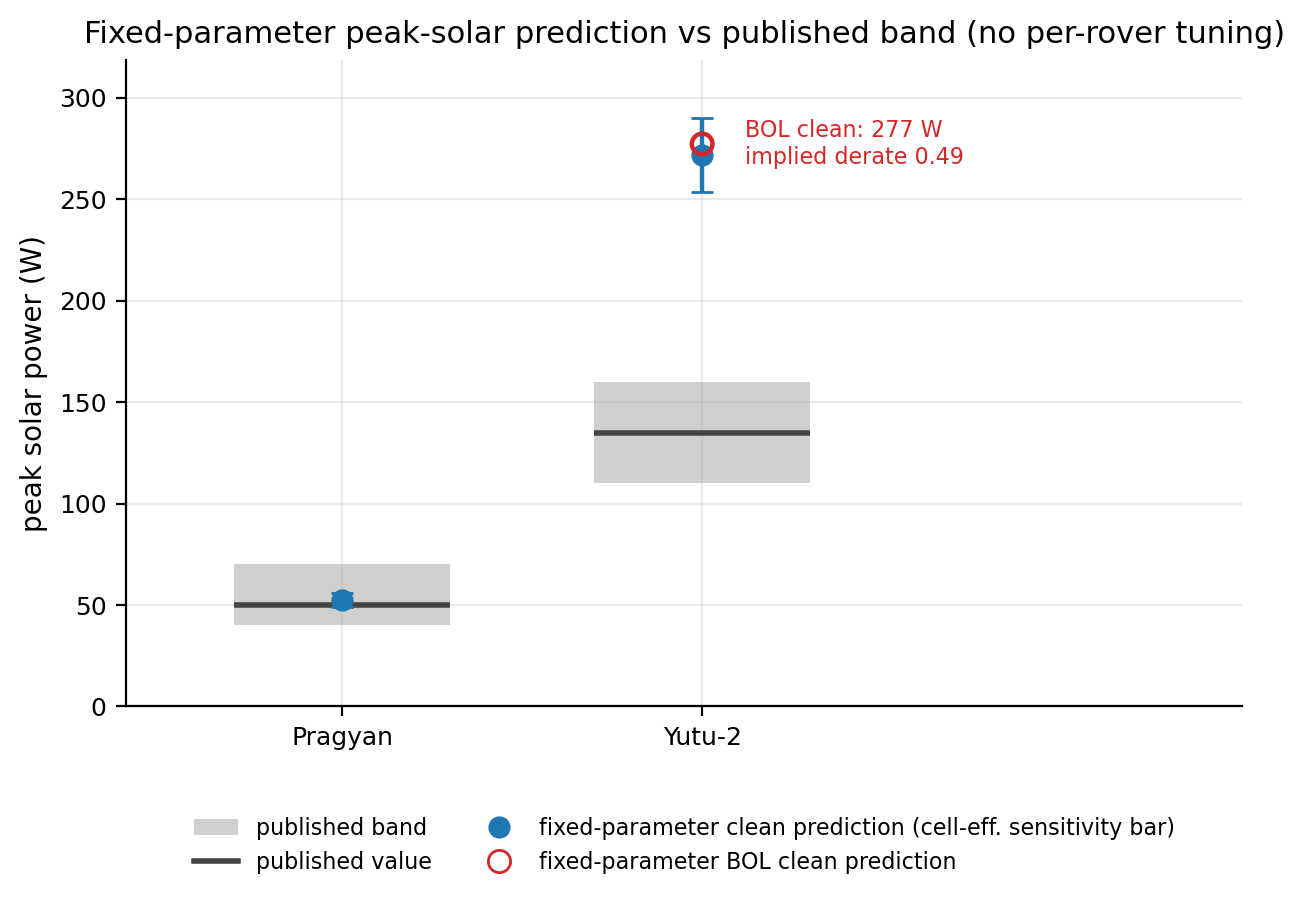}
  \caption{Peak solar prediction versus published flight values.
           Filled markers: model prediction with common beginning-of-life,
           clean-array factors and a cell-efficiency bar ($0.28$--$0.32$).
           Gray band and bar: published number (near-beginning-of-life for
           Pragyan; multi-year flight for Yutu-2). The open marker is that
           same clean prediction on Yutu-2, where $277$~W versus the
           published $135$~W is clean versus dusted, not a like-for-like
           error.}
  \label{fig:peak-solar}
\end{figure}

The result is shown in Fig.~\ref{fig:peak-solar}. Pragyan flew one lunar day, so its published $50$~W (band $40$--$70$~W) is a near-beginning-of-life peak, the same condition the model uses. The prediction is $52$~W ($+5\%$) and stays in-band across the $0.28$--$0.32$ cell-efficiency sweep ($49$--$56$~W). Yutu-2's published $135$~W is a multi-year flight figure, after dust and cell aging, while the model predicts a clean-array $277$~W. The ratio of those two numbers (published / clean $\approx 0.49$) is therefore an implied derate, not a model error on a clean array, and is consistent with the dust accumulation and cell degradation reported for that mission. 

\subsection{Rediscovery check}
\label{sec:rediscovery}
The final evidence layer is a system-level check on the integrated tool: does the optimizer's feasible area actually contain the real rover designs? Due to the small sample size of accessible published micro-rover class designs ($n=5$), we consider this as a sanity check rather than a statistical validation and report it relative to a random baseline. The five rovers used here (Tenacious, CADRE, Rashid-1, Pragyan, and MoonRanger) are those for which a full design vector and a lunar mission scenario could be reconstructed. 

For each rover we run NSGA-II under its associated mission scenario and measure the normalized $L^2$ distance, in the unit-scaled design space, from the real rover design point to the nearest Pareto design point. Of the 11 design variables, the distance uses the eight continuous coordinates the evaluator consumes. Wheelbase is excluded because no sub-model reads it (Section~\ref{sec:evaluator}), so the optimizer has no signal on that coordinate and scoring agreement on it would measure search noise rather than design agreement. Grouser count and mobility architecture are discrete and are left out of the Euclidean metric. This distance is compared for each rover relative to two different baselines. The first is the mean separation between two random points in the unit cube ($\approx1.13$). The second is the same mean, but taken only over randomly drawn designs that are feasible for a rover: they climb the scenario slope, finish with a non-negative energy balance, and travel a nonzero range. That feasible region fills $77$--$92\,\%$ of the box, so the second baseline is $\approx1.10$, only marginally tighter.

\begin{figure}[htbp]
  \centering
  \includegraphics[width=\linewidth]{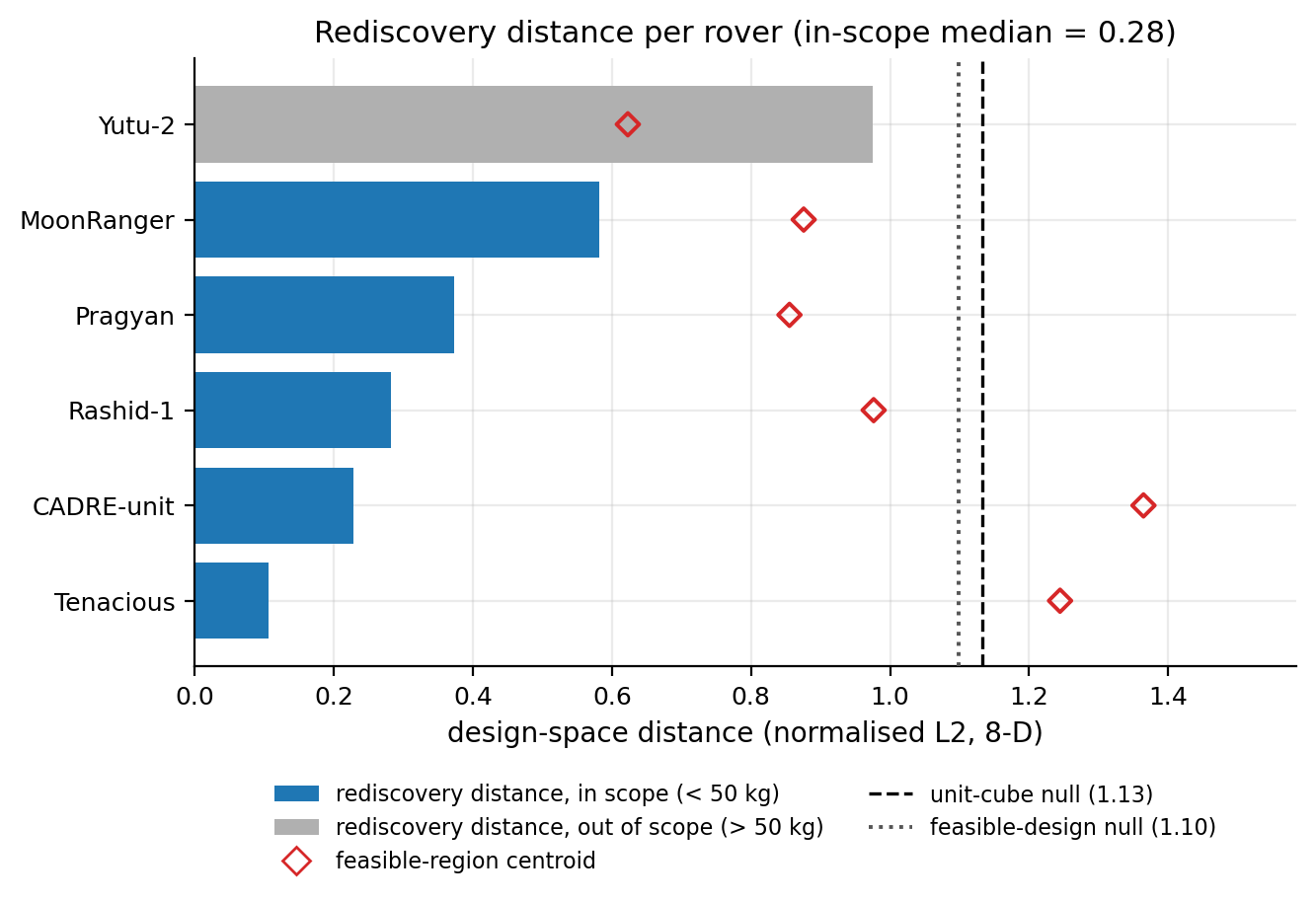}
  \caption{Rediscovery design space distance per rover, colored by
           scope (in-scope $<50$\,kg vs.\ out-of-scope Yutu-2). Red diamonds mark each rover's distance to the feasible-region centroid (the ``typical'' feasible design); the dashed and dotted lines are the random-pair null and the feasible-space null. Every in-scope rover lands closer to its real
           design than the typical feasible design does, while out-of-scope Yutu-2 does not.}
  \label{fig:rediscovery-distance}
\end{figure}

Across the five in-class rovers the median design-space distance is $0.28$, about $25\,\%$ of the random-pair baseline, with every in-scope rover between $9$ and $51\,\%$ of that baseline (Tenacious $0.11$, CADRE unit $0.23$, Rashid-1 $0.28$, Pragyan $0.37$, MoonRanger $0.58$; Fig.~\ref{fig:rediscovery-distance}). Each in-scope rover also lands closer to its real design than the typical feasible design does: its distance falls below its distance to the feasible-region centroid. By comparison, the out-of-scope Yutu-2 sits at $0.97$, or $86\,\%$ of the random-pair baseline and well outside the in-class band, implying the optimizer does not correctly rediscover a rover well beyond the mass range of the micro-rover class. 

The optimizer's front Pareto-dominates five of the six rovers when they are evaluated under the same mission assumptions; this reflects mission constraints the conceptual model does not impose such as radiation hardening, deployability, redundancy, and integration margins, rather than an implication that flown designs are sub-optimal.

\section{Results and discussion}
\label{sec:discussion}

\subsection{Pareto fronts across four canonical scenarios}
\label{sec:pareto}
We apply NSGA-II with the physics-based evaluator as the fitness function across
four canonical lunar scenarios: an equatorial mare traverse, a polar prospecting mission, a crater-rim survey mission, and a highland slope-capability challenge. Each mission scenario is defined by its latitude, soil class, sun geometry, and solar panel orientation (Section~\ref{sec:design-vector}, Table~\ref{tab:design-ranges}). For the three traverse-driven scenarios NSGA-II maximizes range and slope capability while minimizing total mass. The highland slope scenario is posed differently, for the reason given below. The resulting fronts (Fig.~\ref{fig:pareto}) differ sharply in the trade each scenario imposes.

For the three traverse-driven scenarios the slope-capability objective saturates near $33$--$35^\circ$, well above expected operational requirements for those missions, and so the key trade becomes range vs.\ mass. There the fronts span an order of magnitude in range at near-constant slope: the equatorial mare traverse reaches the widest envelope ($\approx0.9$--$80$\,km over $9.6$--$55$\,kg), the crater-rim survey mission spans $\approx0.4$--$25$\,km ($9$--$55$\,kg), and the polar mission front, constrained by the low mean sun elevation and the resulting steep panel tilt required, spans $\approx3.8$--$30$\,km ($12$--$47$\,kg) and has a higher mass floor reflecting the larger battery and solar array required. The polar front reflects only the latitude-driven sun-angle penalty and does not represent terrain horizon masking or site-specific shadowing, which at a real polar site would further constrain the energy budget.

The highland case is formulated as a constrained range--mass trade rather than as a three-objective range--mass--slope trade. On loose regolith, slope capability is dominated by grouser engagement and approaches a common ceiling near $19.6^\circ$, and maximizing it directly therefore collapses the front. We instead require a $15^\circ$ slope capability aligned to the mission scenario and report the resulting range--mass envelope: range rises from $\approx48$\,km at $8$\,kg to a $\approx56$\,km plateau by $\approx20$\,kg, while all Pareto designs retain $18.8$--$19.5^\circ$ of slope capability. This contrast confirms that the dominant trade is set by the mission scenario.

\begin{figure}[htbp]
  \centering
  \includegraphics[width=\linewidth]{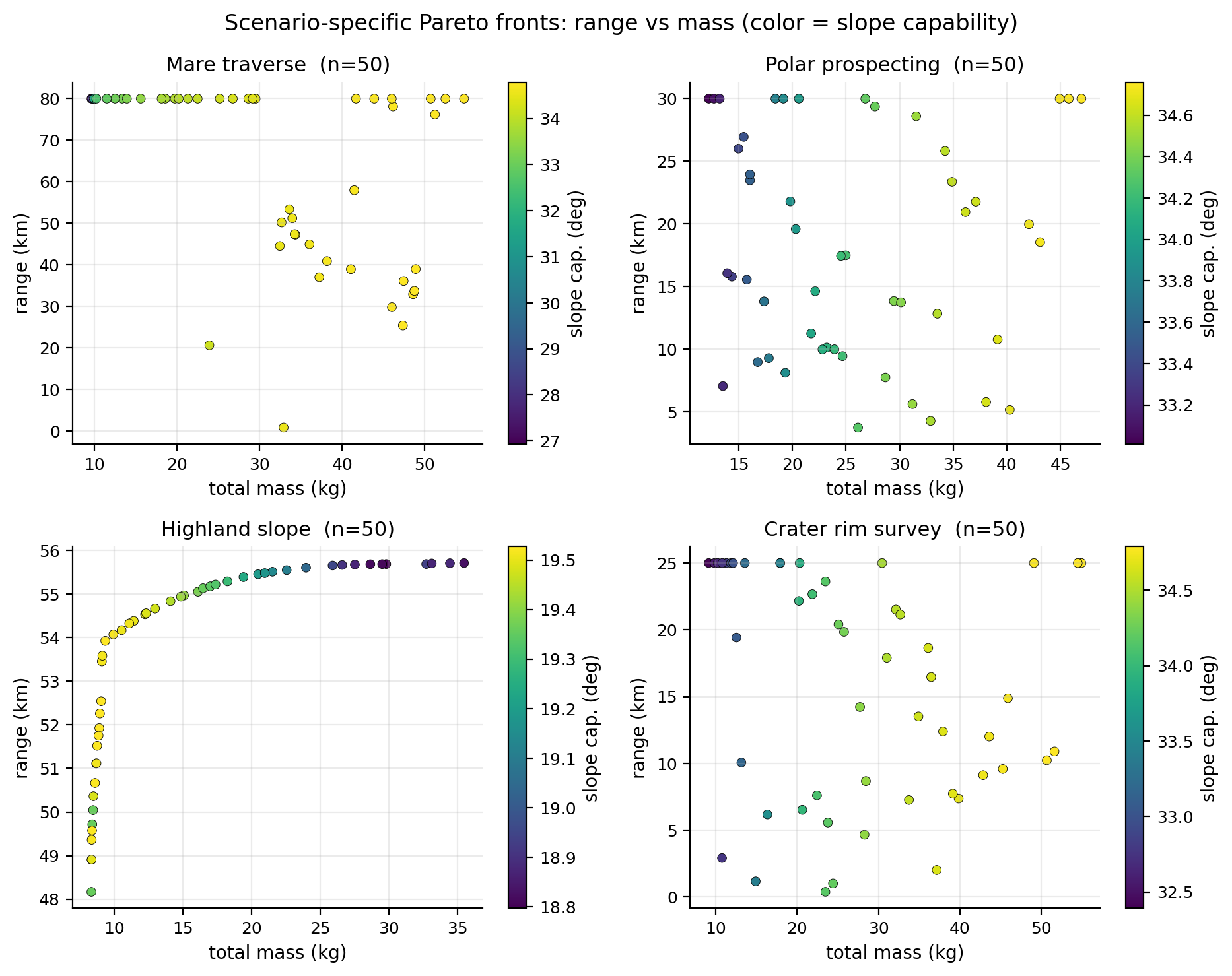}
  \caption{Pareto fronts for the four canonical scenarios (mare, polar,
           highland, crater rim): range vs.\ total mass, colored by slope
           capability.}
  \label{fig:pareto}
\end{figure}

\subsection{Cross-scenario design rules}
\label{sec:design-rules}
Analysis of the design variables of the Pareto-optimal sets exposes a small set of bounded design rules. These are physics-motivated patterns that hold within the modeled objective set and identify which trades remain active once the mission scenario is fixed. The unifying observation is that the limiting objective changes with mission profile even though the mass class is held fixed: the equatorial mare and crater-rim fronts are limited by range, the highland front by slope traction on loose regolith, and the polar front by energy storage. 

\begin{figure}[htbp]
  \centering
  \includegraphics[width=0.8\linewidth]{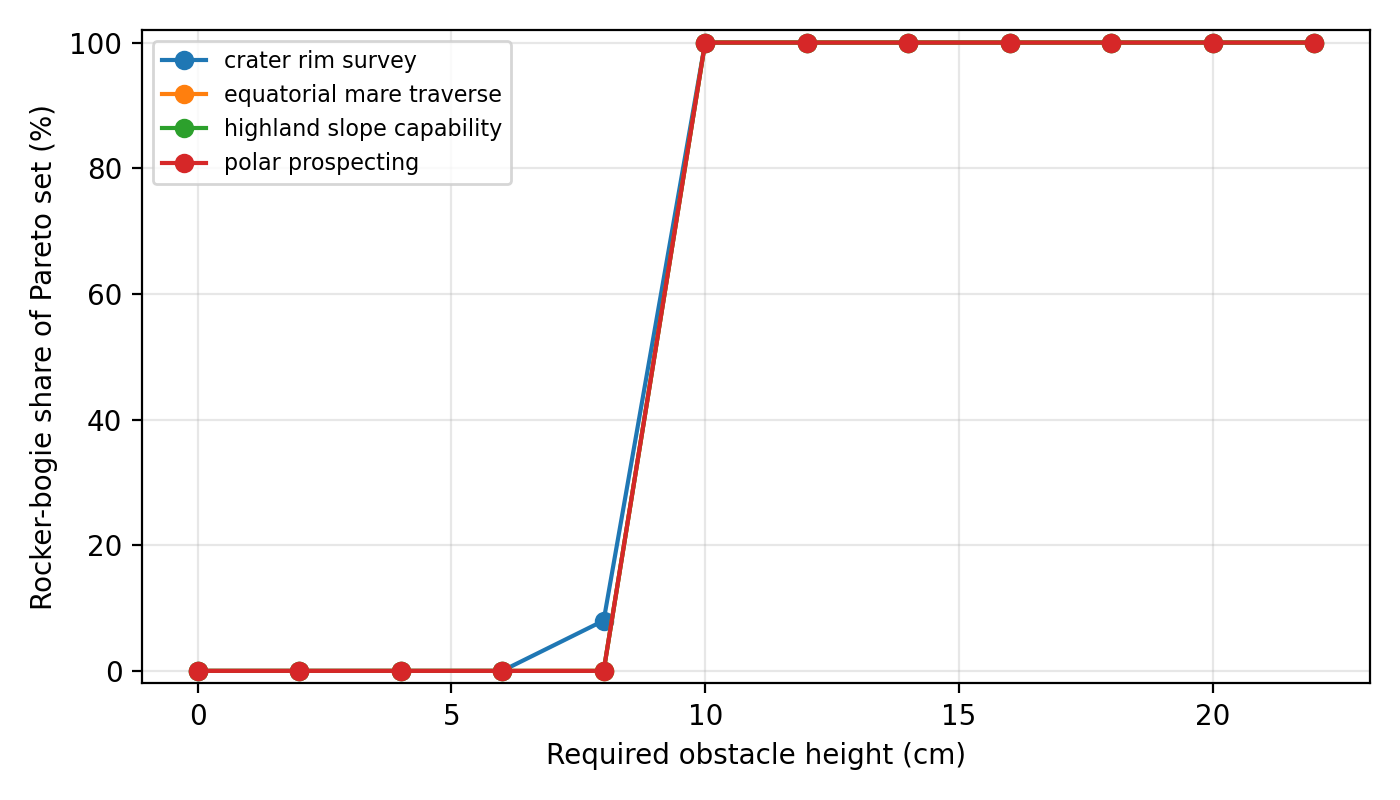}
  \caption{Rocker-bogie share of the Pareto set versus required obstacle
           height across the four canonical scenarios (evaluator-backed
           NSGA-II sweep in the released artifact). Rigid four-wheel
           layouts dominate when $h_{\mathrm{req}}=0$; rocker-bogie
           occupies the front once the requirement exceeds the rigid
           capability at the optimized wheel radius.}
  \label{fig:architecture-crossover}
\end{figure}

\emph{Four wheels dominate the modeled range--mass--slope objectives.} In every scenario and across the entire mass range, Pareto-optimal designs have four wheels when no obstacle navigation requirement is imposed; the extra structural and power cost of two additional drive assemblies outweighs their marginal traction benefit on homogeneous regolith. This overturns our pre-study expectation, drawn from rocker-bogie practice, that the rocker-bogie six-wheel option would become Pareto-optimal above roughly $15$\,kg as per-wheel load and sinkage grew. The fronts instead retain four wheels even at the heavy end because the optimizer can enlarge radius and grousers instead of adding wheels: that geometry recovers the lost traction at far lower mass than two extra drive assemblies. Once a mission requires traversing obstacles taller than the rigid-architecture capability ($\approx 0.5r$), rocker-bogie six-wheel layouts enter the Pareto set (Section~\ref{sec:limitations}, Fig.~\ref{fig:architecture-crossover}). Six-wheel architectures may also be selected for other reasons such as fault tolerance.

\emph{The optimizer maximizes radius and grousers for traction; width falls to the search floor.} The fronts drive wheel radius to its $0.20$\,m ceiling and grousers to their maximum height ($0.020$\,m) and count ($24$), because both are mass-cheap ways to raise modeled drawbar pull. Wheel width goes the other way, and does so in every scenario: all $200$ Pareto points across the four fronts fall in the lowest $5\,\%$ of the searched $0.03$--$0.20$\,m range, and $80$--$100\,\%$ of each front sits on the $0.03$\,m floor itself. That floor is a mass result, not a traction result. The Bekker--Wong kernel does penalize narrow contact patches through higher sinkage and compaction resistance, but once radius and grousers saturate, traction no longer limits the design, and the only term width still affects is wheel mass ($\propto rb$). Width therefore drops until the search bound stops it, so $0.03$\,m marks where we drew that bound rather than an optimum the physics locates. Published in-class micro-rovers use $40$--$80$\,mm contact widths, so a design team applying the tool should set width from the packaging, stability, and bearing-contact constraints that this evaluator does not model (Section~\ref{sec:limitations}).

\emph{At high latitude, the optimizer tilts the panel and grows the battery rather than the array (terrain-free illumination).} In the fixed-tilt approximation, the polar front holds solar-array area near its lower bound (median $0.18$\,m$^2$) while using a steep, pole-facing orientation ($80^\circ$ tilt) to improve incidence given the low sun angle. The remaining energy trade is then carried primarily by battery capacity (median $\approx150$\,Wh, versus $\approx15$\,Wh on the equatorial mare front), which buffers the long low-sun periods. This pattern is the optimizer's response to the latitude-driven sun-angle penalty alone; because the illumination model omits terrain horizon masking and site shadowing, it should be considered as an illustrative consequence of high-latitude geometry rather than a sizing rule for a specific polar site, where local topography may dominate both the orientation and the storage trade.

For a design team applying the tool to a new mission scenario, these patterns suggest a practical order of attack: default to rigid four-wheel layouts unless the mission imposes an obstacle traverse requirement above $\approx 0.5r$, size radius and grousers to meet the scenario's traction requirement, set panel orientation from latitude before sizing the array (revisiting orientation and storage against site-specific terrain at high latitude), and then trade battery capacity and chassis mass against range while setting wheel width from packaging and stability constraints not represented in the objective set.

\subsection{Robustness of the design rules to terramechanics model-form error}
\label{sec:terra-sensitivity}
We next check whether the Pareto fronts and design rules of Section~\ref{sec:pareto} survive error in the wheel--soil model. The test scales the mobilized shear stress $\tau$ inside the contact integrals, which shifts drawbar pull, torque, and sinkage together. Two pairs of multipliers on $\tau$ are used. Factors of $0.85$ and $1.15$ shift median net drawbar pull by $\approx$28\,\% on the digitized single-wheel points, matching the $24$--$27\,\%$ comparison-error band; factors of $0.70$ and $1.30$ shift it by $\approx$55\,\%, a $2\times$ stress envelope. At each scale we re-run the same four-scenario NSGA-II search and re-extract the fronts and design-rule statistics.

\begin{figure}[htbp]
  \centering
  \includegraphics[width=\linewidth]{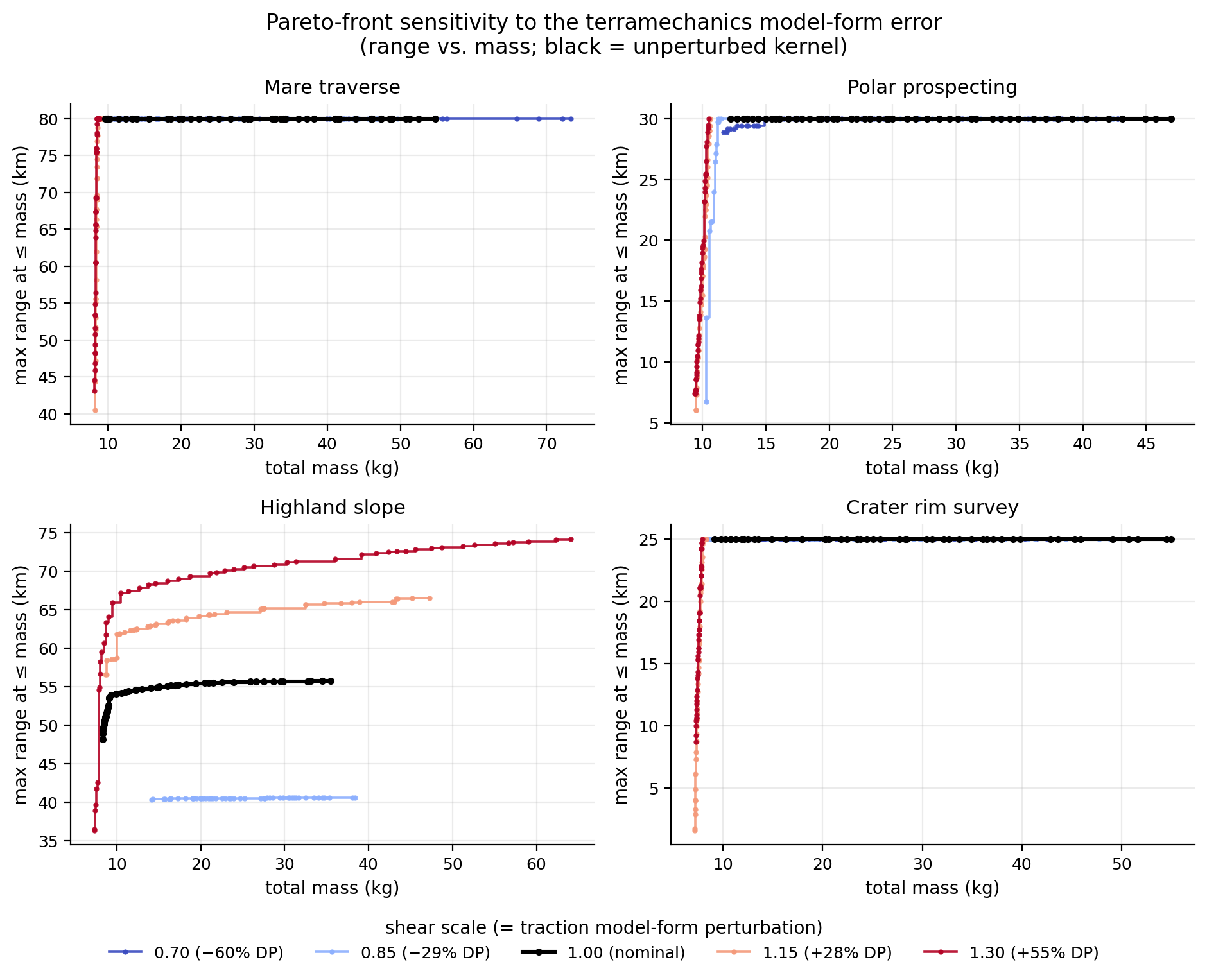}
  \caption{Range-vs.-mass attainment frontiers at five shear-stress scales
           (black = unperturbed kernel; blue/red = traction-pessimistic/
           optimistic, annotated with the induced median drawbar-pull shift).
           Traverse-driven fronts stay pinned to their traverse caps; the
           Highland front shifts with traction and is omitted at the $-55\,\%$
           stress case where the $15^\circ$ slope floor is infeasible.}
  \label{fig:terra-sensitivity}
\end{figure}

Fig.~\ref{fig:terra-sensitivity} and Table~\ref{tab:terra-sensitivity} summarize the sweep, run under the same smooth-regolith, no-obstacle objectives as Section~\ref{sec:pareto}. Across the measured-error band the qualitative design rules hold: every Pareto point remains four-wheeled, the traverse-driven fronts stay at their $80$/$30$/$25$\,km range caps, and slope capability stays at $30$--$35^\circ$ against a $\le25^\circ$ requirement.

If traction is as poor as the nominal kernel, or worse ($\tau$ scale $\le 1$), the optimizer still drives radius and grousers to their ceilings. It backs off those ceilings only when we raise $\tau$ and the soil is easier than the model. Polar energy storage follows the same split: battery capacity stays $\approx10\times$ the mare value and the array stays near its floor unless traction is raised. Those two findings are therefore a conservative design. They still hold if we have overstated how difficult the soil is, but they are not the smallest radius, grouser set, or battery that would suffice if the soil is easier than modeled.

This sweep does not test the width finding. Width is already at its $0.03$\,m lower bound to save mass, not because of traction, so changing $\tau$ leaves it there.

The Highland scenario is the case that does move. Instead of a fixed traverse distance it must meet a $15^\circ$ slope floor, so attained range tracks how much traction the kernel supplies: $41$\,km when drawbar pull is reduced by $28\,\%$, $56$\,km at nominal, $66$\,km when drawbar pull is raised by $28\,\%$, and no feasible design when it is reduced by $55\,\%$. The radius/grouser ceilings, the polar battery premium, and this highland range are therefore the conclusions that depend most on the accuracy of the wheel--soil model.

\begin{table}[htbp]
  \centering
  \small
  \setlength{\tabcolsep}{4pt}
  \caption{Robustness to a shear-stress perturbation over the
           measured $\pm\sim$27\,\% drawbar-pull band ($0.85$/$1.15$) and a
           $2\times$ envelope ($0.70$/$1.30$).}
  \label{tab:terra-sensitivity}
  \begin{tabular}{@{}>{\raggedright\arraybackslash}p{0.27\linewidth}>{\raggedright\arraybackslash}p{0.17\linewidth}>{\raggedright\arraybackslash}p{0.48\linewidth}@{}}
    \toprule
    Conclusion & Verdict & Evidence across the sweep \\
    \midrule
    Four wheels dominate &
      Invariant &
      $100\,\%$ four-wheel Pareto points at every scale (no-obstacle objectives) \\[2pt]
    Range-frontier shape &
      Invariant &
      Traverse caps unchanged; Highland plateau shifts smoothly \\[2pt]
    Slope capability $>$ operational need &
      Holds in band &
      $30$--$35^\circ$ over $0.85$--$1.15$; $\approx24^\circ$ at $2\times$ \\[2pt]
    Radius and grousers peg &
      Conservative &
      Peg at scale $\le1.0$; relaxes under optimistic traction \\[2pt]
    Width at search bound &
      Not transferable &
      Mass-driven bound peg when traction is saturated; flown rovers use $40$--$80$\,mm \\
    Polar storage premium &
      Conservative &
      $\approx10\times$ polar/mare battery for scale $\le1.0$; fades if traction is optimistic \\[2pt]
    Highland range under $15^\circ$ floor &
      Sensitive &
      $41\!\to\!56\!\to\!66$\,km over $\pm28\,\%$; infeasible at $-55\,\%$ \\
    \bottomrule
  \end{tabular}
\end{table}

\subsection{Interpretation of the rediscovery check}
\label{sec:dominance-paradox}
The rediscovery check examines whether the optimizer's feasible area contains real micro-rover designs under class-generic scenarios; it does not estimate mission optimality, and with five in-scope rovers to evaluate against it should not be taken as a statistical validation. The positive result that all in-scope rovers lie much closer to the Pareto set than to a random design baseline is therefore evidence that the integrated model localizes real designs to a plausible region of the design space but should not be taken as an indication that the front is a complete representation of mission decision making.

The Pareto fronts produced by the optimizer dominate most of the published rover points, which can be attributed to the scope of the objectives and inputs. RoverDevKit optimizes range, mass, and slope capability under explicit payload requirements and it does not consider radiation design, deployment mechanisms, obstacle negotiation, fault tolerance, redundancy, communications geometry, assembly, integration, and test (AIT) margin, or science value beyond simple payload mass and power. A flown design can therefore logically be close to the modeled Pareto neighborhood while still sitting behind the mathematical front: the offset from the front is the cost of requirements that the evaluator does not factor in.

This offset can be interpreted as a useful diagnostic: its presence across evaluated flown rovers is consistent with unmodeled robustness margins; for example, the polar fronts show that an idealized panel
orientation can improve the modeled energy budget relative to designs constrained by packaging or deployment requirements. However, showing that the offset is structured and tracks to a specific design factor would require a larger rover registry than is presently available and is therefore left as future work.

\subsection{Limitations}
\label{sec:limitations}
\paragraph{Evaluator fidelity}
The evaluator is intended for conceptual trades, not route-level mission design. The traverse simulation assumes constant slope and homogeneous terrain, with no route planning, spatial obstacle fields, or avoidance behavior; obstacle effects enter only through the architecture-level height requirement discussed below. The single-node thermal model is checked against published survival outcomes rather than flight telemetry. The polar illumination model captures the latitude-driven sun-angle penalty but omits horizon masking and local shadowing, so the polar fronts isolate high-latitude geometry rather than site-specific sizing. The bottom-up mass model carries a mass-dependent bias, over-predicting light rovers and under-predicting heavy ones (Section~\ref{sec:mass-validation}); it is checked against published totals only, so a matching total does not confirm the subsystem allocation. Within the class this compresses the mass axis, so the fronts understate the mass a given range gain costs even though the direction of the trade is unaffected. The ranking in Sections~\ref{sec:runtime} and~\ref{sec:terra-validation} is among these first-order models, not a general claim that locomotion dominates other disciplines at equal fidelity: a thermal or power model built to the same depth as a numerical contact model would likely change both the error ranking and the runtime split. 

\paragraph{Mobility and architecture scope}
The terramechanics kernel is a rigid-wheel, steady-state Bekker--Wong model. It matches drawbar pull within the expected error band on the two primary measured datasets, but under-predicts slip-dependent sinkage and depends on calibrated soil parameters. Section~\ref{sec:terra-sensitivity} propagates this error through the optimizer: the main cross-scenario design rules survive in the conservative direction, while the radius/grouser margin, polar storage premium, and highland range plateau remain the conclusions most contingent on kernel fidelity.

The optimizer also lacks tip-over stability, packaging, and minimum bearing-contact constraints, so the $0.03$\,m wheel-width bound reached in non-traction-limited cases should not be read as a flotation recommendation. Wheelbase is included in the design vector but not used by any sub-model, because load distribution and tip-over stability are not modeled, so the optimizer returns no guidance on it. Adding those two effects is the most direct extension of the present model.

The mobility architecture likewise encodes only first-order effects: rigid four-wheel and rocker-bogie six-wheel layouts differ by obstacle capability and by a suspension mass penalty. Smooth-regolith range--mass--slope objectives therefore favor rigid four-wheel designs when traction is non-binding; rocker-bogie enters the Pareto set only when a mission imposes an obstacle-height requirement (Fig.~\ref{fig:architecture-crossover}). Obstacle negotiation, deformable wheels, dynamic maneuvers, and spatially varying terrain contacts would require a discretized-terrain or physics-based numerical contact model of the kind discussed in Section~\ref{sec:background-theory}.

\paragraph{Validation sample size}
The rediscovery analysis uses only five in-scope micro-rovers with uneven public specifications. It should therefore be interpreted as a sanity check: the reported distances support neighborhood proximity to real designs, but not statistical validation or claims of optimality.

\subsection{Generalization to new regimes and capabilities}
\label{sec:generality}
The present implementation is intentionally scoped to lunar micro-rovers in the $<50$\,kg class. That bound describes the vehicle class rather than a cliff in the physics, and the two ends of the band are set by different things. At the low end the mass model's fixed-overhead terms grow as a fraction of vehicle mass: they are a quarter of published mass at Tenacious's $5$\,kg (Section~\ref{sec:mass-validation}). At the high end the sizing equations remain serviceable---Yutu-2 and MARSOKHOD are predicted to within $9$--$13\,\%$---and what prevents extrapolation is instead the study setup: the design-variable box of Table~\ref{tab:design-ranges}, the restriction to the two mobility architectures modeled here, and the class-generic mission scenarios. The rediscovery check reflects exactly this, localizing every in-class rover while leaving the $135$\,kg Yutu-2 near the random-pair baseline and well outside the in-class band (Section~\ref{sec:rediscovery}). Within the class, the same evaluator can be used for different lunar mission cases by changing mission scenario inputs such as latitude, soil properties, illumination, payload mass and power, duty cycle, and required slope. The design rules reported above should not be transferred directly to heavier rovers, crewed mobility systems, or rovers for other gravity regimes.

Extending the RoverDevKit framework is straightforward but would require new empirical support. Other gravity regimes, especially Mars, would change the wheel-load and power/thermal balances and should be paired with appropriate soil parameters and experimental wheel data. Other design configurations such as six-wheel rocker-bogie architectures could be modeled with better fidelity by replacing the present obstacle-clearance proxy with explicit suspension kinematics, fault-tolerance modeling, and route-dependent obstacle constraints. Finally, replacing the constant-slope capability envelope with terrain-path simulation would move the tool toward route-aware traverse planning while preserving the same evaluator--optimizer workflow.

\section{Conclusions and future work}
\label{sec:conclusions}
This paper presented RoverDevKit, an open evaluator that couples a closed-form wheel--soil model with bottom-up mass, power, thermal-survival, and traverse models and is fast enough to run inside a multi-objective optimizer.

Across mare, polar, highland, and crater-rim missions the limiting factor changes: energy storage at high latitude, slope traction on loose highland regolith, and traverse range on mare and crater-rim terrain. Four-wheel layouts dominate those objectives when no obstacle requirement is imposed. The wheel--soil kernel matches measured single-wheel drawbar pull to within the $20$--$30\,\%$ error typical of this model class, and the mass model predicts published $5$--$50$\,kg rover masses with $10.5\,\%$ median absolute error. Five published in-class designs lie near the designs the optimizer selects, closer to the Pareto front than random feasible points. 

Future work should expand the registry of public rover data toward a sample large enough to strengthen the rediscovery analysis and test whether the remaining offset to flown designs tracks to structured robustness margins such as redundancy, mission duration, or thermal/radiation environment constraints. The traverse simulation should be extended from constant-slope, homogeneous terrain to a route-aware path, so that the four-wheel versus rocker-bogie trade is decided by spatially varying obstacles rather than by the present height-proxy and suspension-mass penalty. For the polar scenario specifically, the smooth-sphere illumination model should be replaced with a terrain-resolved horizon/shadowing model driven by a digital elevation model so that the high-latitude energy budget reflects site topography rather than latitude alone. A higher-fidelity terramechanics model combining SCM/DEM simulation with additional single-wheel testbed data would improve treatment of deformable wheels, obstacle contacts, and soil-parameter uncertainty. The incorporation of explicit operational scheduling across drive/charge/science modes, a broader payload library, and cost, manufacturability and integration constraints would further improve the tool's usefulness for early mission studies.

\section*{Data and code availability}
The RoverDevKit source code, validation data, and paper figure-generation scripts are available in the public GitHub repository at \url{https://github.com/Space-Infrastructure-Lab/roverdevkit}. The repository includes the analytical evaluator, validation datasets, report artifacts, and the interactive browser-based tradespace exploration tool. RoverDevKit is implemented in Python~$3.12$. The evaluator itself depends only on NumPy, SciPy, and PyYAML; multi-objective optimization uses pymoo \cite{blankPymooMultiObjectiveOptimization2020}. The repository includes an environment specification listing all dependencies, and the exact versions used for the results reported here are given in Section~\ref{sec:runtime}. The exact version used for this paper is archived on Zenodo as release \texttt{v1.1.0} (DOI: \href{https://doi.org/10.5281/zenodo.22771969}{10.5281/zenodo.20754999}). All
released versions are available via the concept DOI \href{https://doi.org/10.5281/zenodo.20754998}{10.5281/zenodo.20754998}, which always resolves to the latest release.

\section*{Funding}
This research received no specific grant from any funding agency in the public, commercial, or not-for-profit sectors.

\section*{CRediT authorship contribution statement}
Jon Reifschneider: Conceptualization, Methodology, Software, Validation, Formal
analysis, Investigation, Data curation, Visualization, Writing -- original
draft, Writing -- review \& editing.

\section*{Declaration of competing interest}
The author declares that they have no known competing financial interests or personal relationships that could have appeared to influence the work reported in this paper.

\section*{Declaration of AI use in the manuscript preparation process}
During the preparation of this work, the author used Claude Opus 4.8 to assist with software development and during review to obtain suggestions for improving the clarity and organization of the manuscript. After using this tool, the author reviewed the suggestions and edited the content as needed and takes full responsibility for the content of the published article.


\bibliographystyle{elsarticle-num}
\bibliography{main}

\end{document}